\documentclass[11pt]{article}

\usepackage[final]{acl}

\usepackage{times}
\usepackage{latexsym}

\usepackage[T1]{fontenc}

\usepackage[utf8]{inputenc}

\usepackage{makecell}
\usepackage{amsmath}
\usepackage{amssymb}
\usepackage[table]{xcolor} 
\usepackage[most]{tcolorbox}
\definecolor{badred}{RGB}{200, 40, 40}
\definecolor{goodgreen}{RGB}{40, 160, 40}
\definecolor{bg_case}{RGB}{248, 249, 250} 
\definecolor{line_case}{RGB}{220, 220, 220}
\definecolor{academicblue}{RGB}{235, 245, 255}
\usepackage{tikz}
\usetikzlibrary{shapes, arrows.meta, positioning, calc, backgrounds, fit, decorations.pathreplacing}

\usepackage{pgfplots}
\pgfplotsset{compat=1.18}

\usepackage{microtype}

\usepackage{inconsolata}

\usepackage{graphicx}

\usepackage{booktabs}
\usepackage{multirow}
\usepackage{algorithm}
\usepackage{algpseudocode} 
\usepackage{amsmath}
\usepackage{listings}  
\usepackage{tcolorbox}
\tcbuselibrary{skins, breakable}

\usepackage{enumitem}

\author{
Yixin Bu$^{1,2,*}$, Runze Xia$^{1,2,*}$, Guanyun Zou$^{1,2}$, 
Yupeng Ji$^{3}$, Haodong Liu$^{3}$, Piji Li$^{1,2,^\dagger}$\\
\textsuperscript{\rm 1}College of Artificial Intelligence,\\
Nanjing University of Aeronautics and Astronautics, China\\
\textsuperscript{\rm 2}The Key Laboratory of Brain-Machine Intelligence Technology,\\
Ministry of Education, Nanjing, China\\
\textsuperscript{\rm 3}Researcher\\
\texttt{\{sx2416068,xiarunze,zouguanyun,pjli\}@nuaa.edu.cn, \{yupengji,liuhaodong\}@163.com}
}

\title{DUD: Decoupled Update Dynamics for Reliable Uncertainty Quantification in Large Language Models}
\begin{document}
\maketitle

\renewcommand{\thefootnote}{\fnsymbol{footnote}}
\footnotetext[2]{Corresponding author.}
\renewcommand{\thefootnote}{\arabic{footnote}}
\begin{abstract}
Accurate Uncertainty Quantification (UQ) is critical for reliable deployment of Large Language Models (LLMs), yet traditional probability-based metrics often fail to capture the model's true epistemic state. While recent mechanistic approaches leverage hidden state dynamics, they typically aggregate residual stream updates, conflating the distinct roles of parametric memory (Feed-Forward Networks) and contextual processing (Attention). We argue that this aggregation obscures fine-grained mechanistic conflicts, such as memory-context misalignment, that are fundamental indicators of uncertainty. To address this, we introduce \textbf{D}ecoupled \textbf{U}pdate \textbf{D}ynamics \textbf{(DUD)}, a framework that explicitly decouples FFN and Attention contributions via noise-induced causal interventions. By quantifying the independent restoration capabilities of each module, we construct a dual-stream dynamic profile that captures the model's internal fragility. Extensive experiments demonstrate that DUD significantly outperforms state-of-the-art baselines in both uncertainty estimation and calibration, while exhibiting superior cross-dataset generalization, validating decoupled dynamics as a robust proxy for model faithfulness.
\end{abstract}

\section{Introduction}
\label{sec:intro}

Reliable Uncertainty Quantification (UQ) is a prerequisite for deploying Large Language Models (LLMs) in high-stakes environments. Ideally, a model's predicted confidence should perfectly align with the correctness of its generation. However, achieving such calibration remains a formidable challenge. Traditional UQ approaches, ranging from black-box sampling consistency~\cite{manakul2023selfcheckgpt} to white-box logit analysis~\cite{kuhn2023semantic}, predominantly rely on the final output distribution. While effective for capturing data-level ambiguity (aleatoric uncertainty), these surface-level metrics often fail to reflect the model's true epistemic state. By treating the inference process as a black box and focusing solely on the end result, they overlook the internal computational dynamics where uncertainty actually originates. Consequently, a model may output a high probability score while its internal reasoning process is fraught with instability, leading to severe miscalibration.

To bridge this gap, Mechanistic Interpretability offers a fundamental theoretical grounding. Pioneering studies have mapped functional specialization within Transformer architectures, identifying Feed-Forward Networks (FFNs) as repositories for parametric knowledge~\citep{meng2022locating} and Multi-Head Self-Attention (MHSA) as routers for contextual information~\citep{elhage2021mathematical}. These insights have been actively adapted in related fields, particularly in hallucination detection, where researchers utilize hidden state trajectories to pinpoint generation errors~\citep{chen2024inside, marks2023geometry,zhang2025icr}. However, we identify a critical methodological oversight in how these mechanistic signals are currently utilized for UQ: existing approaches typically employ aggregation strategies, merging the contributions of memory and context into unified representations. We argue that this aggregation obscures the essence of uncertainty. In complex reasoning, epistemic uncertainty often arises precisely from the mechanistic conflict between these components such as when parametric memory contradicts contextual cues. By aggregating these opposing signals, current methods neutralize the fine-grained evidence required for precise diagnosis.

In this paper, we propose Decoupled Update Dynamics (DUD), a framework that transforms UQ from passive estimation to active mechanistic diagnosis. DUD explicitly disentangles the causal contributions of FFNs and MHSA modules by measuring their independent ability to restore prediction confidence from a noise-induced perturbed state. This decoupled tracing reveals a critical phenomenon overlooked by previous studies: uncertainty manifests as distinct spatiotemporal fragility patterns. We observe that generation failures are not uniform but stem from specific mechanistic breakdowns---manifesting as routing instability in early-layer MHSA during context encoding, and a precipitous collapse in late-layer FFNs during knowledge retrieval. By capturing these fine-grained conflicts, we construct a dual-stream dynamic profile that serves as a robust proxy for the model's epistemic state. Extensive experiments demonstrate that DUD outperforms state-of-the-art baselines in both uncertainty estimation and calibration. Notably, our method exhibits exceptional cross-dataset generalization, confirming that these mechanistic signatures are intrinsic to the model's reasoning process rather than artifacts of specific tasks.

Our contributions are summarized as follows:

\begin{itemize}[leftmargin=*, nosep]

    \item \textbf{Decoupled Causal Framework:} We introduce the first UQ framework that explicitly disentangles the causal contributions of FFNs and MHSA modules, demonstrating that internal module dynamics provide a significantly more faithful proxy for reliability than traditional output probabilities.

    \item \textbf{Mechanistic Fragility Discovery:} Through rigorous causal tracing, we uncover that prediction uncertainty physically manifests as systemic fragility and module conflict, particularly within middle-layer FFNs, offering a physical grounding for detecting model errors.

    \item \textbf{Empirical Performance:} Extensive experiments across diverse benchmarks demonstrate that our decoupled approach significantly outperforms both logits-based baselines and aggregated mechanistic probes, achieving superior calibration and error detection capabilities.

\end{itemize}
\section{Related Work}
\label{sec:related_work}

\subsection{Uncertainty Quantification in LLMs}

Traditional uncertainty quantification predominantly relies on output probabilities, such as Semantic Entropy \cite{kuhn2023semantic}. However, these logit-based metrics remain superficial, assessing final confidence without inspecting the internal reasoning process, often failing to distinguish inherent data ambiguity from model capability failures. Recent approaches leverage intermediate representations to address this. Methods like SAPLMA \cite{chen2024inside}, Lookback Lens \cite{chuang2024lookback}, and DoLa \cite{chuang2023dola} utilize hidden states, attention weights, or layer-wise contrasts. While richer than logits, these methods typically aggregate signals from different modules into unified representations. This conflation obscures fine-grained internal conflicts---specifically between parametric memory and contextual cues that are critical indicators of generation failure.

\subsection{Mechanistic Interpretability of Transformer Components}

Mechanistic interpretability research establishes that Multi-Head Self-Attention (MHSA) and Feed-Forward Networks (FFN) perform distinct computational roles: MHSA acts as a contextual signal router transferring information \cite{elhage2021mathematical}, while FFNs serve as key-value memories for parametric knowledge \cite{geva2021transformer, meng2022locating}. \citet{stolfo2023mechanistic} further demonstrated their dominance at different processing stages of inference. Building on these insights, our work diverges from previous aggregation-based UQ methods. We explicitly model the dynamic evolution of these decoupled streams, quantifying uncertainty as the mechanistic misalignment between parametric memory (FFN) and contextual processing (MHSA).

\section{Methodology}
\label{sec:methodology}

\begin{figure*}[t]
    \centering
    \includegraphics[width=1\textwidth]{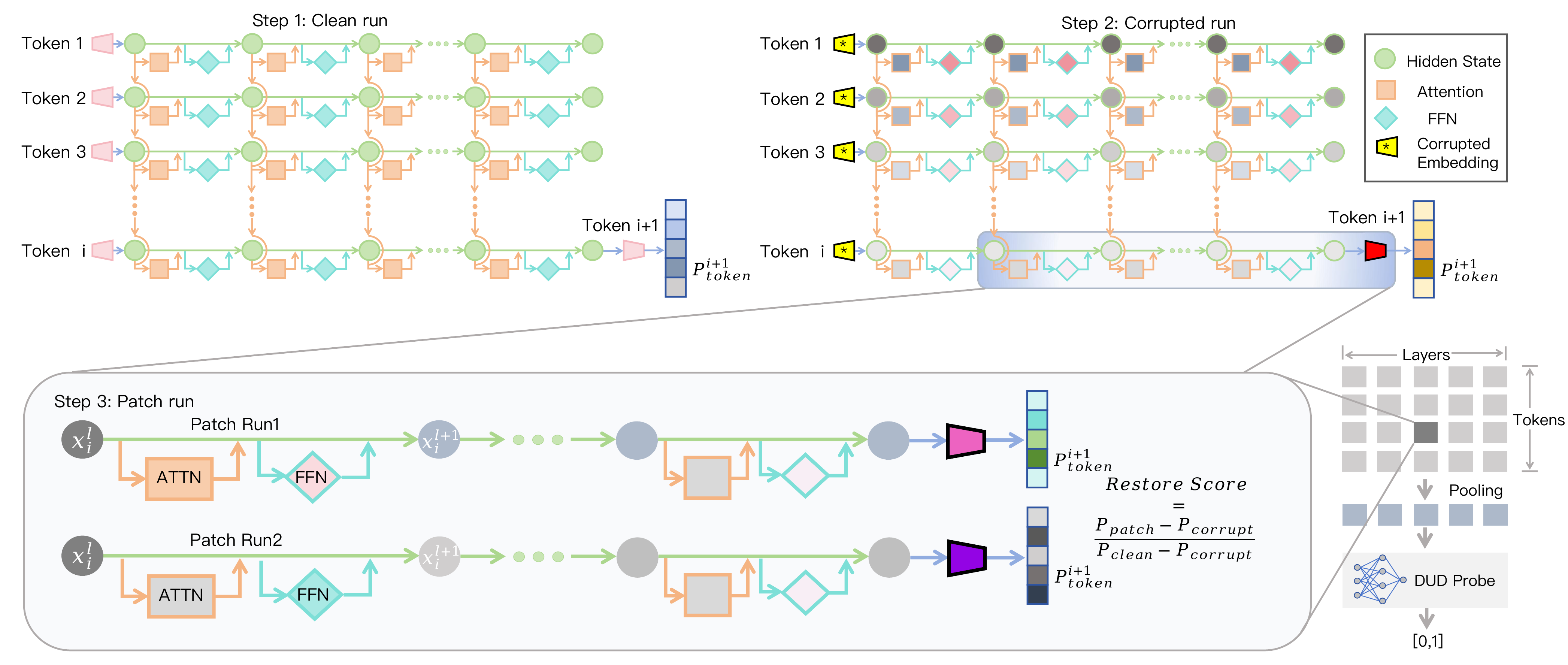}
    \caption{
        Overview of the \textbf{Decoupled Update Dynamics (DUD)} framework. 
        The process involves three stages: 
        (1) \textbf{Clean Run:} A standard forward pass to cache hidden states; 
        (2) \textbf{Corrupted Run:} Noise is injected into embeddings to induce an information bottleneck; 
        (3) \textbf{Patch Run:} We perform causal interventions by independently restoring the activations of \textcolor{gray}{Attention} and \textcolor{pink}{FFN} modules layer-by-layer in Step 2, followed by recovering \textcolor{orange}{Attention} and \textcolor[RGB]{0,255,255}{FFN} in Step 3. 
        The resulting probability shifts are normalized into restoration scores, forming a $2L$-dimensional dynamic profile that trains the DUD-Probe to detect uncertainty.
    }
    \vspace{-1em}
    \label{fig:framework}
\end{figure*}

We propose the Decoupled Update Dynamics (DUD) framework, a mechanistic approach designed to quantify model uncertainty by dissecting the internal tug-of-war between parametric memory and contextual processing. As illustrated in Figure~\ref{fig:framework}, our method transforms uncertainty quantification from a passive observation task into an active causal interrogation process.

\subsection{Preliminaries: Transformer Dynamics}
Given an input sequence $x = [x_1, \dots, x_T]$, the hidden state $\mathbf{h}_i^{(l)}$ at layer $l$ and token position $i$ is updated as the following illustrative logic:
\begin{equation}
    \mathbf{h}_i^{(l)} = \mathbf{h}_i^{(l-1)} + \mathbf{a}_i^{(l)} + \mathbf{m}_i^{(l)}
\end{equation}
where $\mathbf{a}_i^{(l)}$ is the output of the Multi-Head Self-Attention (MHSA) module, responsible for contextual routing; and $\mathbf{m}_i^{(l)}$ is the output of the Feed-Forward Network (FFN), which serves as a key-value memory for parametric knowledge \citep{geva2021transformer}. The final probability distribution for the next token is obtained by projecting the final layer state $\mathbf{h}_T^{(L)}$ through the unembedding matrix $\mathbf{E}$ and applying a softmax function:
\begin{equation}
    P(y | x) = \text{Softmax}(\mathbf{E} \mathbf{h}_T^{(L)})
\end{equation}

\subsection{Decoupled Causal Tracing}
\label{subsec:decoupled_tracing}

In this subsection, we introduce a decoupled causal tracing framework to isolate and quantify the causal contributions of memory- and context-related components under uncertainty. Traditional uncertainty metrics break down when memory and context provide conflicting evidence.
We explicitly model this mismatch. Our method applies decoupled causal tracing under three execution settings: Clean, Corrupted, and Patch runs. Comparing these runs reveals how memory and context individually influence the final prediction (Figure~\ref{fig:framework}). 

\paragraph{Establishing Baselines via Clean and Corrupted Runs.}
We first define the upper and lower bounds of the model's performance to quantify the information bottleneck.
\begin{itemize}[leftmargin=*, nosep]
    \item \textbf{Clean Run:} We perform a standard forward pass with the original input $x$. We cache the clean activations (hidden states $\mathbf{h}$, MHSA outputs $\mathbf{a}$, and FFN outputs $\mathbf{m}$) for all layers and record the prediction probability $P_{\text{clean}}(y_t)$.
    \item \textbf{Corrupted Run:} We inject Gaussian noise $\epsilon \sim \mathcal{N}(0, \nu)$ into the input embeddings to simulate a loss of context ($\mathbf{e}^*(x) = \mathbf{e}(x) + \epsilon$). This degrades the prediction probability to $P_{\text{corr}}(y_t)$.
\end{itemize}
The probabilistic gap between these two runs represents the Total Effect (TE) of the noise, calculated as the average drop across the sequence:
\begin{equation}
    \text{TE} = \frac{1}{T} \sum_{t=1}^{T} \left[ P_{\text{clean}}(y_t) - P_{\text{corr}}(y_t) \right]
\end{equation}

\paragraph{Decoupled Patch Run.}
To disentangle the roles of the two pathways, we perform Component-Specific Interventions. Unlike standard causal tracing which restores the entire hidden state, we execute separate restored runs for the FFN and Attention modules.
For a specific layer $l$, we intervene in the corrupted forward pass by replacing the activation of a target module $\phi \in \{\text{MHSA}, \text{FFN}\}$ with its value cached from the Clean Run, while keeping all other components in the Corrupted state.
Mathematically, restoring module $\phi$ at layer $l$ yields a restored probability $P_{\text{restored}}^{(l, \phi)}(y_t)$. This operation isolates the causal contribution of that specific module in recovering the correct prediction from the noisy baseline.

\paragraph{Sequence-Level Dynamic Profiling.}
We quantify the contribution of each module by calculating the Restoration Score. The score $\mathcal{S}_{\phi}^{(l)}$ represents the normalized probability shift induced by the Restored Run:
\begin{equation}
    \mathcal{S}_{\phi}^{(l)} = \frac{1}{T} \sum_{t=1}^{T} \frac{P_{\text{restored}}^{(l, \phi)}(y_t) - P_{\text{corr}}(y_t)}{\text{TE}}
\end{equation}
Empirically, we observe that $\mathcal{S}_{\phi}^{(l)}$ is predominantly negative, as restoring a single module within a corrupted stream introduces state incoherence. We therefore interpret the \textit{magnitude} of this suppression as a proxy for uncertainty. Specifically, a small magnitude ($\mathcal{S} \approx 0$) indicates mechanistic robustness, where the model tolerates the internal mismatch, suggesting the generation is grounded in stable features. In contrast, a large negative magnitude ($|\mathcal{S}| \gg 0$) signals mechanistic fragility, where single-module restoration causes a catastrophic drop in probability, revealing that the generation relies on a precarious internal equilibrium.

\subsection{DUD-Probe Construction}
\label{subsec:probe}

To translate these fine-grained internal dynamics into a comprehensive uncertainty estimate, we construct a lightweight, non-linear probe.

\textbf{Feature Engineering.}
We concatenate the layer-wise restoration scores into a unified feature vector $\mathbf{v} \in \mathbb{R}^{2L}$, applying Z-score normalization to ensure numerical stability:
\begin{equation}
    \mathbf{v} = \text{Norm}\left( \text{Concat}\left[ \mathcal{S}_{\text{attn}}^{(1..L)}, \mathcal{S}_{\text{ffn}}^{(1..L)} \right] \right)
\end{equation}
This vector preserves the structural relationship between memory and context streams.

\textbf{Probe Design and Training.}
We formulate uncertainty quantification as a binary classification task. Ground truth labels are derived from the ROUGE-L score ($\mathcal{S}_{\text{rouge}}$) of the generated response:
\begin{equation}
    y_i = 
    \begin{cases} 
    0 \ (\text{Correct}) & \text{if } \mathcal{S}_{\text{rouge}} \ge \tau \\
    1 \ (\text{Uncertain}) & \text{otherwise}
    \end{cases}
\end{equation}
Unlike linear classifiers, we employ a 4-layer MLP ($2L\!\rightarrow\!128\!\rightarrow\!64\!\rightarrow\!32\!\rightarrow\!1$)
 to capture non-linear interactions. We incorporate LeakyReLU, Batch Normalization to handle negative restoration scores and prevent overfitting. The probe is trained using Binary Cross-Entropy (BCE) loss to predict generation errors.

\section{Empirical Preliminary Analysis}

In our Decoupled Update Dynamics (DUD) framework, the core objective of the decoupled restoration score is to quantify model uncertainty, specifically by assessing its ability to distinguish between correct and incorrect predictions, and to serve as a robust signal for guiding probe training. To this end, we conduct a series of experiments across four distinct cognitive task datasets: HaluEval and SQuAD (context-rich), and TriviaQA and HotpotQA (context-free). These experiments validate the effectiveness of the proposed decoupled restoration scores in distinguishing correct and incorrect predictions, and demonstrate their potential in providing valuable signals for effective probe training.

\begin{figure}[t]
    \centering
    \includegraphics[width=0.48\textwidth]{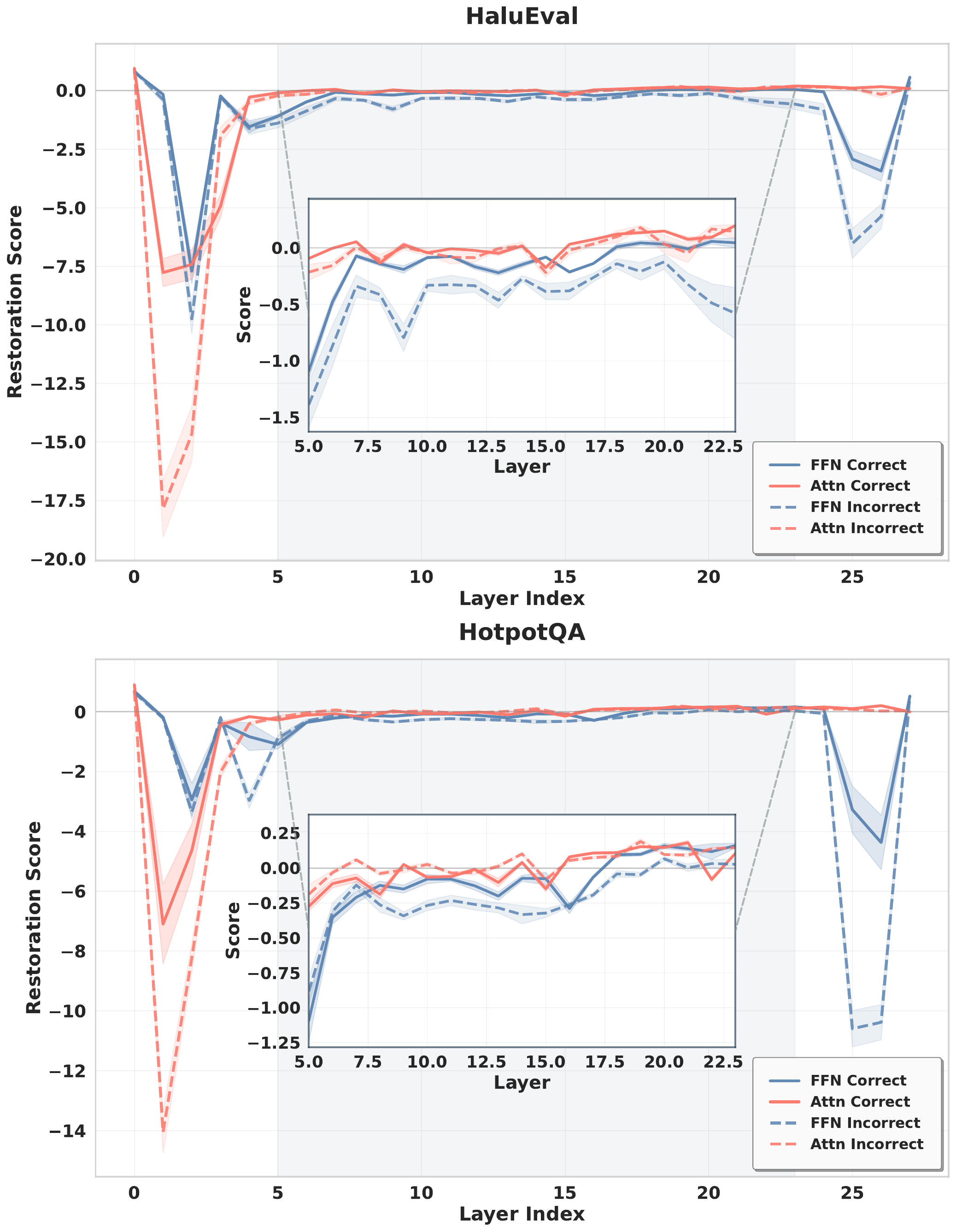}
    \caption{Layer-wise restoration dynamics on HaluEval (Top) and HotpotQA (Bottom). The curves show the restoration scores for correct and incorrect predictions across layers.}
    \label{fig:main_causal_tracing}
    \vspace{-1em}
\end{figure}

\subsection{Restoration Dynamics: Fragility and Task Dependence}
\label{subsec:restoration_dynamics}

Figure~\ref{fig:main_causal_tracing} visualizes layer-wise restoration dynamics, revealing a distinct mechanistic signature of uncertainty. Across all datasets, we observe a consistent pattern in the restoration scores. In the \textit{early layers (0-5)}, both Attention and FFN modules exhibit significant fluctuations, with restoration scores showing larger differences between correct and incorrect predictions. This indicates a high degree of model uncertainty during the initial encoding phase, where errors are often driven by Attention failures, particularly in contextual routing. In contrast, the \textit{middle layers} demonstrate relative stability, with restoration scores exhibiting smaller variations, which suggests a more stable integration of information as the model processes the input. However, a critical divergence emerges in the \textit{late layers (20-28)}, where we observe a specific collapse in FFN scores while Attention remains stable. The sharp decline in FFN restoration scores highlights a fundamental failure in parametric memory during the later stages of generation, marking this as a key indicator of model uncertainty. These observed differences in the restoration scores across layers serve as crucial signals for guiding probe training, allowing for the effective identification of model uncertainty.

The magnitude of these patterns varies with task demands. In context-rich tasks like HaluEval, early layers show greater instability due to the challenge of processing long contexts. In contrast, HotpotQA, a knowledge-intensive task, exhibits particularly large changes in the late-layer FFN scores. This could be due to the model's reliance on internal synthesis for answering complex questions, as it lacks external context to ground its reasoning. As a result, late-layer FFN stability becomes more crucial, with large score shifts indicating the model's difficulty in maintaining internal coherence without sufficient external information.

\begin{figure}[t]
    \centering
    \includegraphics[width=0.48\textwidth]{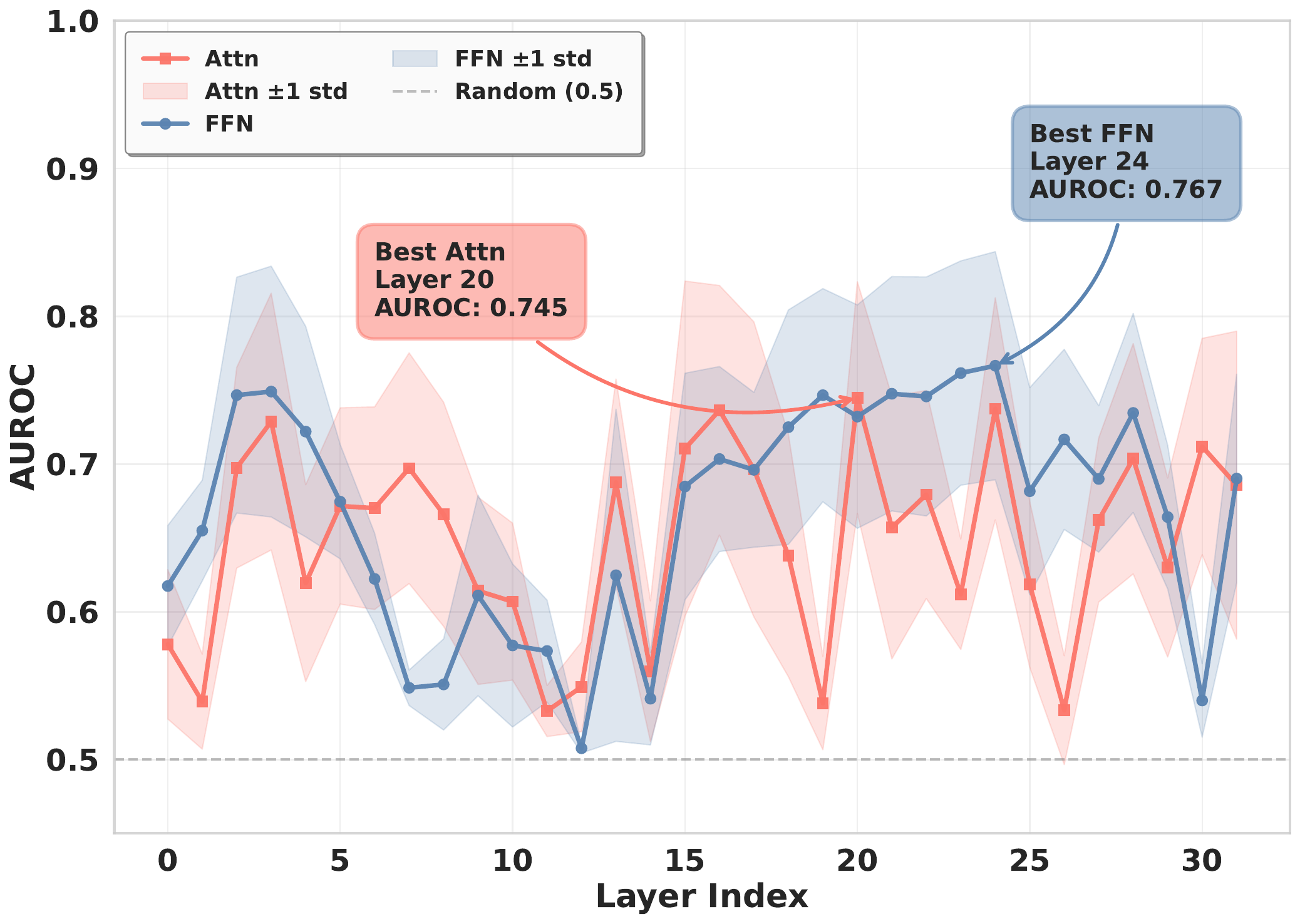}
    \caption{Layer-wise AUROC analysis for LLaMA-3.1-8B, averaged across four datasets. The FFN (blue) and Attention (red) curves demonstrate that restoration scores from both modules effectively distinguish uncertainty across the network depth.}
    \vspace{-1em}
    \label{fig:layer_auroc}
\end{figure}

\subsection{Discriminative Power: AUROC Analysis}
\label{subsec:auroc_analysis}

To assess whether decoupled restoration scores can reliably distinguish correct from incorrect generations and serve as effective signals for probe learning, Figure~\ref{fig:layer_auroc} shows that these scores exhibit strong and structured discriminative power across network depth. For LLaMA-3.1-8B, elevated AUROC values consistently emerge in the early layers (e.g., layers 2-3) as well as in the middle-to-late layers (approximately layers 16-24), closely mirroring the layer-wise mechanistic deviations identified in the restoration dynamics analysis. This alignment indicates that the proposed scores capture predictable and functionally meaningful uncertainty signals tied to distinct stages of representation processing, rather than arbitrary fluctuations. As a result, the restoration scores form informative and structured features for uncertainty probing, enabling reliable separation between correct and incorrect generations. Similar AUROC profiles are observed across Qwen-2.5-7B and Gemma-2-9B (Appendix~\ref{app:layer_auroc_analysis}), suggesting that the discriminative capacity of the decoupled restoration scores generalizes across model architectures.

\subsection{The Confidence Paradox: Mechanistic Deviations vs. Surface Certainty}
\label{subsec:quadrant_analysis}

\begin{figure}[t]
    \centering
    \includegraphics[width=1.0\columnwidth]{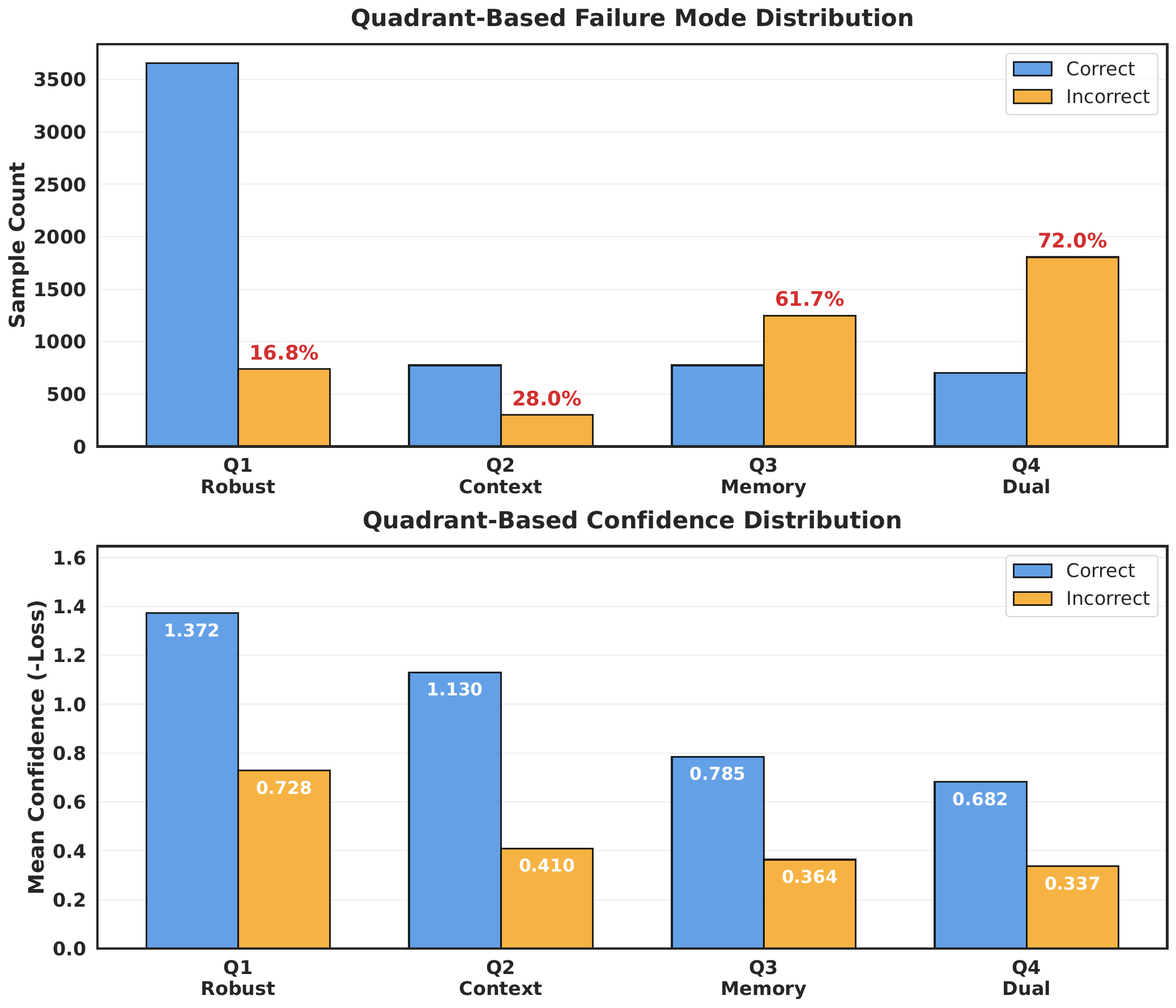}
    \caption{\textbf{Quadrant Analysis on SQuAD (Layer 24).} \textbf{Top:} Error rates across four mechanistic quadrants defined by FFN and Attention deviation. \textbf{Bottom:} Confidence score distributions for Correct vs. Incorrect predictions within each quadrant, highlighting the misalignment between surface confidence and internal mechanistic stability.}
    \label{fig:quadrant_analysis}
    \vspace{-1em}
\end{figure}

To further examine why surface-level confidence often fails to reflect generation reliability, we conduct a quadrant analysis on SQuAD at Layer 24. By establishing deviation thresholds at the 75th percentile of correct predictions, we partition samples into four mechanistic regimes based on the stability of Attention and FFN modules. These regimes span from Q1 (Robust), a stable state where both modules remain within nominal limits, through localized instability in Q2 (Context Deviant) where only the Attention score exceeds the threshold and Q3 (Memory Deviant) where only the FFN score exceeds it, to Q4 (Dual Deviant), which signals systemic fragility with both modules violating the stability bounds.

The results (Figure~\ref{fig:quadrant_analysis}) reveal a clear misalignment between internal stability and logits-based confidence. Error rates surge with accumulated deviation, rising from 16.8\% in the robust Q1 quadrant to 72.0\% in the dual-deviant Q4, confirming that internal instability is a decisive predictor of failure. Crucially, surface confidence fails to track this risk: incorrect predictions in stable Q1 often retain deceptively high confidence (Stubborn Errors), while correct predictions in deviant Q4 suffer from low confidence (Humble Truths). This ``Confidence Inversion'' demonstrates that probability thresholds alone are insufficient. By grounding prediction risk in decoupled mechanistic states, DUD-Probe effectively identifies high-risk generations that surface confidence overlooks.

\begin{table*}[t]
\centering

\small
\setlength{\tabcolsep}{3.2pt} 
\renewcommand{\arraystretch}{1.2} 

\begin{tabular}{l|cccc|cccc|cccc}
\toprule
\multirow{2}{*}{\textbf{Methods}} & \multicolumn{4}{c|}{\textbf{Gemma-2-9B}} & \multicolumn{4}{c|}{\textbf{Qwen2.5-7B}} & \multicolumn{4}{c}{\textbf{Llama-3.1-8B}} \\ 
\cmidrule(lr){2-5} \cmidrule(lr){6-9} \cmidrule(lr){10-13}
 & \textbf{Halu} & \textbf{SQ} & \textbf{Hot} & \textbf{Triv} & \textbf{Halu} & \textbf{SQ} & \textbf{Hot} & \textbf{Triv} & \textbf{Halu} & \textbf{SQ} & \textbf{Hot} & \textbf{Triv} \\ 
\midrule

PPL 
& 0.5431 & 0.5415 & 0.7038 & 0.7535  
& 0.5553 & 0.5201 & 0.6032 & 0.6649  
& 0.5720 & 0.6231 & 0.6577 & 0.7238 \\ 

LN-Entropy 
& 0.7531 & 0.6491 & 0.7227 & 0.7388 
& 0.7177 & 0.6342 & 0.6957 & 0.7431 
& 0.6431 & 0.6203 & 0.6643 & 0.5994 \\

Semantic Entropy 
& 0.7745 & 0.7483 & 0.7361 & 0.7482 
& 0.7293 & 0.7001 & 0.7142 & 0.7248 
& 0.7745 & 0.7041 & 0.7143 & 0.7362 \\

SAR 
& 0.7856 & 0.7643 & 0.7217 & 0.7391 
& 0.7647 & 0.7113 & 0.6845 & 0.7432 
& 0.7793 & 0.7531 & 0.7335 & 0.7647 \\

LLM-Check 
& 0.5745 & 0.5703 & 0.5411 & 0.5593 
& 0.5143 & 0.5555 & 0.5431 & 0.5729 
& 0.5348 & 0.5331 & 0.5501 & 0.5275 \\

SAPLMA 
& 0.8100 & 0.7015 & 0.8201 & 0.7755 
& 0.7901 & 0.6832 & 0.7507 & \underline{0.8149} 
& 0.7328 & 0.7007 & 0.7638 & 0.7650 \\

SEP 
& 0.6331 & 0.6645 & 0.6218 & 0.7563 
& 0.6647 & 0.6432 & 0.6507 & 0.7667 
& 0.7547 & 0.7309 & 0.6537 & 0.7241 \\

ICR Probe 
& \underline{0.8358} & \underline{0.8204} & \underline{0.8333} & \underline{0.7917} 
& \underline{0.8144} & 0.7203 & 0.7771 & 0.7544 
& 0.7703 & 0.7548 & \underline{0.8139} & 0.7557 \\

\rowcolor{academicblue} 
\textbf{DUD-Probe (Ours)} 
& \textbf{0.8747} & \textbf{0.8218} & \textbf{0.8597} & \textbf{0.8199}  
& \textbf{0.8642} & \textbf{0.8204} & \textbf{0.8435} & \textbf{0.8252}  
& \textbf{0.9487} & \textbf{0.8959} & \textbf{0.8579} & \textbf{0.8031} \\ 

\bottomrule
\end{tabular}
\caption{Comparison of uncertainty quantification performance (AUROC) across three LLMs and four datasets. By displaying models horizontally, we highlight that \textbf{DUD-Probe} consistently achieves the best performance (bold) across different architectures. Halu: HaluEval, SQ: SQuAD, Hot: HotpotQA, Triv: TriviaQA.}
\label{tab:main_results_wide}
\end{table*}

\section{Experiments}

\subsection{Experimental Setup}
\label{sec:exp_setup}

\paragraph{Models, Datasets, and Baselines}
We evaluate three open-source LLMs Llama-3-8B-Instruct \citep{grattafiori2024llama}, Qwen2.5-7B-Instruct \citep{hui2024qwen2}, and Gemma-2-9B-it \citep{team2024gemma} across four knowledge-intensive benchmarks: HaluEval \citep{li2023halueval} (Question Answering) , SQuAD \citep{rajpurkar2016squad} (Reading Comprehension), TriviaQA \citep{joshi2017triviaqa} (Retrieval), and HotpotQA \citep{yang2018hotpotqa} (Reasoning). 
We benchmark against diverse baselines, including training-free methods (\textbf{PPL} \citep{ren2022out}, \textbf{LN-Entropy} \citep{malinin2020uncertainty}, \textbf{SE} \citep{kuhn2023semantic}, \textbf{SAR} \citep{duan2024shifting}, \textbf{LLM-Check}) and training-based probes (\textbf{SAPLMA} \citep{chen2024inside}, \textbf{SEP} \cite{kossen2024semantic}, and the aggregated mechanistic \textbf{ICR Probe} \cite{zhang2025icr}). Detailed configurations are provided in Appendix~\ref{appendix:experimental}.

\paragraph{Evaluation Protocol and Implementation}
We employ a rigorous two-stage protocol. First, ground truth correctness is determined via ROUGE-L \citep{lin2004rouge}, where generations meeting the semantic threshold are labeled as \textit{Correct} ($y=0$), otherwise \textit{Incorrect} ($y=1$). Second, uncertainty performance is primarily assessed using AUROC \citep{fawcett2006introduction}, with auxiliary metrics (ECE \cite{guo2017calibration}, PRR \cite{malinin2018predictive}) and sensitivity analyses (BERTScore \cite{zhang2019bertscore}, EM \cite{rajpurkar2016squad}) detailed in the Appendix~\ref{app:threshold_analysis}.
For implementation, the DUD-Probe is trained on decoupled features ($\mathbf{v} \in \mathbb{R}^{2L}$) using 5-fold cross-validation, ensuring identical data splits across all baselines for fair comparison.

\subsection{Main Results: Uncertainty Quantification Performance} \label{sec:main_results}

Table~\ref{tab:main_results_wide} demonstrates that DUD-Probe consistently achieves state-of-the-art uncertainty quantification performance across diverse models and datasets. Compared to logits-based Semantic Entropy, DUD-Probe yields substantial improvements (e.g., +17.4\% on LLaMA-3.1 HaluEval), indicating that internal causal dynamics provide substantially richer reliability signals than surface probabilities. Moreover, DUD-Probe outperforms the aggregated ICR Probe (e.g., +5.0\% on Qwen2.5), validating the necessity of decoupling Attention and FFN contributions. By separating these two update pathways, our method exposes fine-grained mechanistic conflicts that are obscured by aggregated representations. This advantage is consistent across architectures (LLaMA, Qwen, Gemma), with near-perfect detection on LLaMA-3.1 HaluEval (AUROC 0.949), suggesting that stronger models exhibit more distinguishable internal signatures of correctness.

\begin{figure*}[!t] \centering \includegraphics[width=1\textwidth]{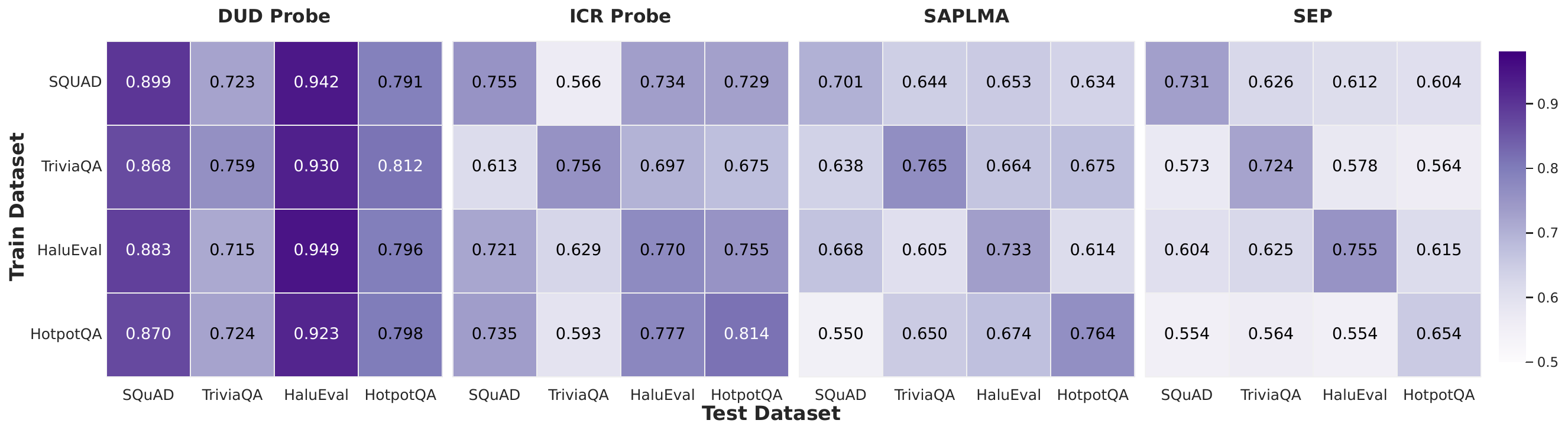} \caption{Cross-dataset generalization heatmaps for ICR Probe, SAPLMA, and SEP. Each subplot displays the AUROC when probe is trained on the row dataset and tested on column dataset, with values annotated in each cell.} \vspace{-1em} \label{fig:generalization} \end{figure*}

To assess cross-dataset transferability, we train DUD-Probe on source datasets and evaluate on unseen targets (Figure~\ref{fig:generalization}). DUD-Probe exhibits markedly stronger generalization than baselines, maintaining high AUROC scores where SAPLMA and SEP frequently degrade below 0.6. For example, transferring from TriviaQA to SQuAD achieves an AUROC of 0.868, substantially surpassing ICR (0.613). Notably, out-of-distribution performance often approaches in-domain results (e.g., TriviaQA $\rightarrow$ HaluEval: 0.930), indicating that DUD-Probe captures task-agnostic mechanistic signatures such as FFN fragility rather than dataset artifacts. Overall, DUD-Probe exhibits minimal degradation ($<5\%$), compared to $15\%\mathbin{-}20\%$ drops in baselines.

We further examine robustness to the inherent ambiguity in defining correctness for open-ended generation by varying the ROUGE-L threshold used for labeling. As shown in Table~\ref{tab:rouge_sensitivity}, DUD-Probe consistently achieves the highest AUROC under Lenient (0.3), Standard (0.5), and Strict (0.7) criteria across all datasets. In contrast, baselines such as SEP and SAPLMA exhibit substantial performance fluctuations. Importantly, under the strictest setting, where generations must closely match references, DUD-Probe maintains strong performance, suggesting that the learned features capture intrinsic correctness rather than superficial lexical overlap.

Finally, to provide a comprehensive assessment, we report additional evaluation results in the Appendix. As detailed in Appendix~\ref{app:additional_metrics}, DUD-Probe consistently improves calibration error (ECE) and rejection efficiency (PRR). We also analyze robustness across model scales (Appendix~\ref{app:model_size}), confirming stable performance from smaller to larger parameter regimes. Sensitivity analysis with respect to alternative labeling metrics (EM and BERTScore) further verifies that the observed gains are not artifacts of a particular evaluation choice (Appendix~\ref{app:threshold_analysis}).

\begin{table}[h]
\centering
\small
\setlength{\tabcolsep}{3.5pt} 
\renewcommand{\arraystretch}{1.1} 
\begin{tabular}{c|c|cccc}
\toprule
\textbf{TH} & \textbf{Method} & \textbf{Halu} & \textbf{SQ} & \textbf{Hot} & \textbf{Triv} \\ \midrule
\multirow{4}{*}{\shortstack{Lenient\\$(>0.3)$}} 
 & SEP & 0.6147 & 0.6493 & 0.6645 & 0.7503 \\
 & SAPLMA & 0.7119 & 0.7135 & 0.7047 & 0.7531 \\
 & ICR Probe & 0.7438 & 0.7001 & 0.7399 & 0.7769 \\
 & \textbf{DUD (Ours)} & \textbf{0.8528} & \textbf{0.7614} & \textbf{0.8579} & \textbf{0.8476} \\ \midrule
\multirow{4}{*}{\shortstack{Standard\\$(>0.5)$}} 
 & SEP & 0.6647 & 0.6432 & 0.6507 & 0.7667 \\
 & SAPLMA & 0.7901 & 0.6832 & 0.7507 & 0.8149 \\
 & ICR Probe & 0.8144 & 0.7203 & 0.7771 & 0.7544 \\
 & \textbf{DUD (Ours)} & \textbf{0.8642} & \textbf{0.8204} & \textbf{0.8435} & \textbf{0.8252} \\ \midrule
\multirow{4}{*}{\shortstack{Strict\\$(>0.7)$}} 
 & SEP & 0.6047 & 0.6214 & 0.6535 & 0.7443 \\
 & SAPLMA & 0.7105 & 0.7031 & 0.7149 & 0.7348 \\
 & ICR Probe & 0.7549 & 0.6943 & 0.7344 & 0.7599 \\
 & \textbf{DUD (Ours)} & \textbf{0.8135} & \textbf{0.8347} & \textbf{0.8636} & \textbf{0.8345} \\ \bottomrule
\end{tabular}
\caption{Robustness analysis on Qwen2.5-7B under varying ROUGE-L thresholds . DUD-Probe consistently outperforms baselines across all settings, demonstrating robustness to labeling criteria.}
\label{tab:rouge_sensitivity}
\vspace{-2em}
\end{table}

\subsection{Mechanistic Ablation Analysis}
\label{sec:ablation}

To dissect the sources of DUD-Probe's performance, we conduct ablation studies on module contributions and layer hierarchy.

\paragraph{Impact of Decoupled Components.} 
A core premise of DUD-Probe is that FFN (parametric memory) and Attention (contextual routing) provide distinct, complementary uncertainty signals. As shown in the upper section of Table~\ref{tab:ablation_combined}, while single-stream probes (\textbf{Attn Only} and \textbf{FFN Only}) significantly outperform the random baseline, the Full dual-stream configuration consistently achieves the highest AUROC across all datasets. 
This confirms that FFN and Attention capture different aspects of generation failure for instance, a factual error might trigger an FFN collapse while leaving Attention relatively stable. By integrating both streams, the DUD-Probe effectively captures the full spectrum of uncertainty, validating the necessity of our decoupling strategy.

\paragraph{Impact of Layer Groups.} We investigate the distribution of uncertainty signals by removing Early (EL, 1-10), Middle (ML, 11-21), and Deep (DL, 22-32) layer groups (Table~\ref{tab:ablation_combined}, Part II). The \textbf{w/o EL} setting causes the most significant drop on HaluEval (
0.95 $\rightarrow$ 0.85), confirming that early-layer instability reflects encoding failures in context-heavy tasks. Conversely, removing Middle and Deep layers universally degrades performance, particularly on reasoning tasks like HotpotQA. These results indicate that there is no single ``golden layer'' for uncertainty detection; instead, the probe relies on the holistic dynamic trajectory from initial encoding to final projection to make reliable judgments.
\begin{table}[h]
\centering
\small
\setlength{\tabcolsep}{4pt} 
\begin{tabular}{@{}l|cccc@{}}
\toprule
\textbf{Configuration} & \textbf{Halu} & \textbf{SQ} & \textbf{Hot} & \textbf{Triv} \\ 
\midrule
\multicolumn{5}{l}{\textit{\textbf{I. Module Contribution}}} \\ 
Random & 0.5000 & 0.5000 & 0.5000 & 0.5000 \\
Attn Only & 0.9144 & 0.8349 & 0.7933 & 0.7385 \\
FFN Only  & 0.8968 & 0.8511 & 0.7872 & 0.7447 \\
\rowcolor{academicblue!30} 
Full &  \textbf{0.9487} & \textbf{0.8959} & \textbf{0.8579} & \textbf{0.8031} \\ 

\midrule
\multicolumn{5}{l}{\textit{\textbf{II. Layer Contribution (removing from Full)}}} \\ 
w/o EL & 0.8511 & 0.8351 & 0.7845 & 0.7164 \\
w/o ML & 0.8893 & 0.8241 & 0.7576 & 0.7281 \\
w/o DL & 0.9001 & 0.8024 & 0.8005 & 0.7542 \\ 
\bottomrule
\end{tabular}
\caption{Mechanistic Ablation Study on Llama-3.1-8B. We analyze the impact of decoupled modules (I) and layer groups (II). Halu: HaluEval, SQ: SQuAD, Hot: HotpotQA, Triv: TriviaQA.}
\label{tab:ablation_combined}
\vspace{-1em}
\end{table}

\subsection{Case Study: Decoupling Faithfulness from Confidence}
\label{subsec:case_study}

To intuitively understand how DUD-Probe operates beyond surface statistics, we present a token-level visualization of two representative cases in Table~\ref{tab:case_study_token}. We define the probe's output as a \textit{Faithfulness Score}, displayed as subscripts.

\textbf{Case 1: The Stubborn Error.} 
The model answers the question about ``stick gymnastics'' with ``Japan''. Crucially, the model's output probability (Logits) is extremely high ($\approx 0.96$), indicating a state of deceptive certainty. However, our DUD-Probe assigns near-zero scores to the generated tokens (e.g., $\text{Japan}_{0.04}$). This confirms that despite the smooth output distribution, the internal mechanistic support for this answer is collapsed. The probe successfully punctures the veil of overconfidence to reveal the hidden epistemic failure.

\textbf{Case 2: The Humble Truth.} 
Conversely, for the ``Land of Fish and Rice'' query, the model's confidence is low ($\approx 0.35$), likely due to the rarity of the entity in the training distribution. A logits-based detector would flag this as uncertain. In contrast, DUD-Probe maintains high faithfulness scores (Avg 0.72) across the tokens. This indicates that the internal retrieval and routing mechanisms are actually stable, correctly identifying the generation as a valid, albeit ``hesitant'' truth.

These cases demonstrate that our decoupled dynamics capture the \textit{intrinsic reliability} of the generation process, which is often effectively orthogonal to the model's calibrated confidence.

\begin{table}[t]
\centering
\small

\begin{tcolorbox}[
    colback=bg_case, colframe=line_case, boxrule=0.8pt, arc=3mm,
    title={\textbf{The "Overconfidence" Trap} \hfill \textit{\footnotesize(Confident but Wrong)}},
    coltitle=black, colbacktitle=gray!15, fonttitle=\small,
    left=2pt, right=2pt, top=2pt, bottom=2pt
]
    
    \begin{tabular}{@{}p{0.08\columnwidth} p{0.86\columnwidth}@{}}
        \textbf{Q:} & What corresponds to the sport of stick gymnastics? \\
    \end{tabular}
    \vspace{-4pt} 
    
    \begin{tabular}{@{}p{0.08\columnwidth} p{0.86\columnwidth}@{}}
        \textbf{A:} & It$_{\textcolor{badred}{0.12}}$ is$_{\textcolor{badred}{0.09}}$ \textbf{Japan}$_{\textcolor{badred}{0.04}}$ .$_{\textcolor{badred}{0.08}}$ \\
    \end{tabular}
    
    \tcbline

    \vspace{-2pt}
    \begin{tabular}{@{}p{0.48\columnwidth} @{}r@{}}
        \textbf{Logits Conf:} \textbf{0.96} (High) & \textbf{DUD Score:} \textcolor{badred}{\textbf{0.08}} (Low) \\
    \end{tabular}
    \textit{\footnotesize $\to$ DUD correctly flags mechanistic collapse.}
\end{tcolorbox}

\vspace{-4pt}

\begin{tcolorbox}[
    colback=bg_case, colframe=line_case, boxrule=0.8pt, arc=3mm,
    title={\textbf{The "Underconfidence" Save} \hfill \textit{\footnotesize(Hesitant but Correct)}},
    coltitle=black, colbacktitle=gray!15, fonttitle=\small,
    left=2pt, right=2pt, top=2pt, bottom=2pt
]
    
    \begin{tabular}{@{}p{0.08\columnwidth} p{0.86\columnwidth}@{}}
        \textbf{Q:} & What is the province traditionally known as? \\
    \end{tabular}
    \vspace{-4pt}
   
    \begin{tabular}{@{}p{0.08\columnwidth} p{0.86\columnwidth}@{}}
        \textbf{A:} & Land$_{\textcolor{goodgreen}{0.65}}$ of$_{\textcolor{goodgreen}{0.72}}$ Fish$_{\textcolor{goodgreen}{0.81}}$ and$_{\textcolor{goodgreen}{0.68}}$ Rice$_{\textcolor{goodgreen}{0.75}}$ \\
    \end{tabular}
    
    \tcbline
    
    \vspace{-2pt}
    \begin{tabular}{@{}p{0.48\columnwidth} @{}r@{}}
        \textbf{Logits Conf:} \textbf{0.35} (Low) & \textbf{DUD Score:} \textcolor{goodgreen}{\textbf{0.72}} (High) \\
    \end{tabular}
    \textit{\footnotesize $\to$ DUD confirms internal robustness.}
\end{tcolorbox}

\caption{Token-level visualization of DUD faithfulness scores (subscripts). DUD effectively penalizes overconfident hallucinations (Case 1) while validating underconfident truths (Case 2).}
\label{tab:case_study_token}
\vspace{-1em}
\end{table}

\section{Conclusion}
We introduced the \textbf{Decoupled Update Dynamics (DUD)} framework, shifting uncertainty quantification to active mechanistic diagnosis. By disentangling FFN and Attention contributions, we revealed that uncertainty manifests as systemic fragility and internal conflict. Our dual-stream probe effectively captures these signals, achieving state-of-the-art performance and robust generalization. This demonstrates that monitoring the dynamic interplay of internal components offers a superior reliability proxy compared to surface-level probabilities or aggregated signals.

\section*{Limitations}

While effective, our approach has limitations. First, as a white-box method requiring access to internal activations and causal interventions, DUD is inapplicable to closed-source API-based models (e.g., GPT-4). Second, although we optimized the probe efficiency, the causal tracing process involves multiple forward passes (clean, corrupted, and patched runs), which increases inference latency compared to simple logits-based methods. Future work will focus on distilling these mechanistic signals into lighter-weight detectors to enable real-time monitoring without heavy computational overhead.

\section*{Acknowledgments}
This research is supported by the National Natural Science Foundation of China (No.62476127),  the Natural Science Foundation of Jiangsu Province (No.BK20242039), the Basic Research Program of the Bureau of Science and Technology (ILF24001), the Scientific Research Starting Foundation of Nanjing University of Aeronautics and Astronautics (No.YQR21022), and the High Performance Computing Platform of Nanjing University of Aeronautics and Astronautics.


\bibliography{custom}

\appendix

\section{Complete Causal Tracing Results}
\label{app:full_causal_tracing}

This section presents the complete layer-wise restoration score analysis across all four evaluation datasets, complementing the main text's focused discussion on representative datasets (HaluEval and HotpotQA).

\begin{figure*}[t]
    \centering
    \includegraphics[width=0.95\textwidth]{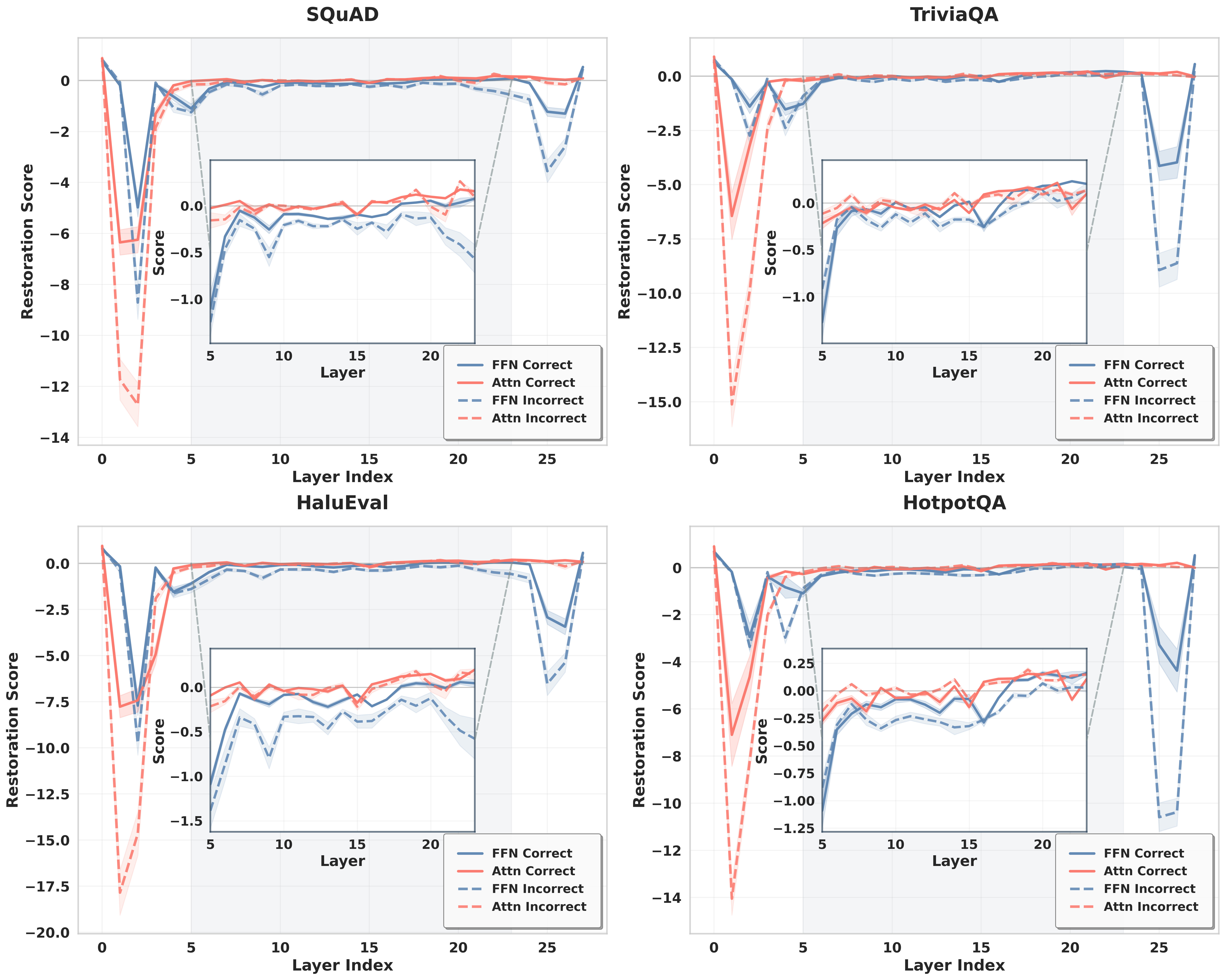}
    \caption{Layer-wise restoration scores across all four datasets: SQuAD (top-left), TriviaQA (top-right), HaluEval (bottom-left), and HotpotQA (bottom-right). Solid lines represent correct predictions; dashed lines represent incorrect predictions. Blue curves: FFN modules; Red curves: Attention modules. Inset plots show zoomed views of middle layers (6-22).}
    \label{fig:full_causal_tracing}
\end{figure*}

Figure~\ref{fig:full_causal_tracing} provides a comprehensive view of the restoration dynamics across SQuAD, TriviaQA, HaluEval, and HotpotQA. Several consistent patterns emerge across all datasets:

\paragraph{Universal Layer-wise Patterns:}
Across all four datasets, we observe the characteristic "double-valley" structure discussed in Section 3, with Attention modules (red curves) exhibiting deeper negative scores in early layers (0-5), while FFN modules (blue curves) dominate in late layers (20-28). This universal pattern validates the temporal specialization of different modules in the prediction pipeline.

\paragraph{Middle-Layer Dynamics:}
The inset plots provide zoomed views of the middle layers (6-22), revealing critical differences between correct and incorrect predictions. In all datasets, correct predictions (solid lines) show restoration scores approaching zero or slightly positive, indicating successful semantic integration. In contrast, incorrect predictions (dashed lines) maintain persistent negative scores (approximately -0.3 to -0.5), suggesting locked-in erroneous trajectories. Notably, FFN scores consistently remain below Attention scores in this region, confirming FFN's dominant role in knowledge retrieval as discussed in Section 4.1.

\paragraph{Task-Specific Variations:}
Comparing context-rich tasks (SQuAD, HaluEval) with context-free tasks (TriviaQA, HotpotQA) reveals the computational burden shift identified in Section 4.2. Context-rich tasks exhibit deeper early-layer FFN valleys (reflecting input encoding costs), while context-free tasks show more pronounced late-layer FFN drops (reflecting parametric memory retrieval demands). These task-dependent dynamics underscore the adaptability of our method across different information sources.

\paragraph{Robustness vs. Fragility:}
The magnitude gap between correct (solid) and incorrect (dashed) predictions is consistent across all datasets, with incorrect predictions showing $2\text{--}3\times$ deeper negative scores. This systematic difference evident in early valleys, middle-layer persistence, and late-layer drops provides the signal our probe leverages for hallucination detection. The consistency of this pattern across diverse tasks (reading comprehension in SQuAD, factual QA in TriviaQA, multi-hop reasoning in HotpotQA, and hallucination-focused evaluation in HaluEval) demonstrates the generalizability of restoration-based uncertainty quantification.

\paragraph{Cross-Dataset Consistency:}
Despite differences in task formulation and difficulty, all four datasets exhibit qualitatively similar restoration dynamics. Standard deviation analysis (not shown) confirms that the temporal ordering of module contributions (early Attention$\to$middle FFN $\to$ late FFN) remains stable across datasets, analogous to the consistency observed in ICR scores (see ICR Probe Figure 5). This stability suggests that restoration scores capture intrinsic model dynamics rather than dataset-specific artifacts, supporting their use as a robust uncertainty signal.

\section{Layer-wise AUROC Analysis Across Models and Datasets}
\label{app:layer_auroc_analysis}

This section provides comprehensive layer-wise AUROC analysis across all three evaluated models (Qwen-2.5-7B, LLaMA-3.1-8B, Gemma-2-9B) and all four datasets (SQuAD, TriviaQA, HaluEval, HotpotQA). Each figure shows the discriminative power of individual layer restoration scores, demonstrating the consistency of temporal patterns across different model architectures and task types.

\subsection{Qwen-2.5-7B-Instruct}

\begin{figure*}[t]
    \centering
    \includegraphics[width=0.95\textwidth]{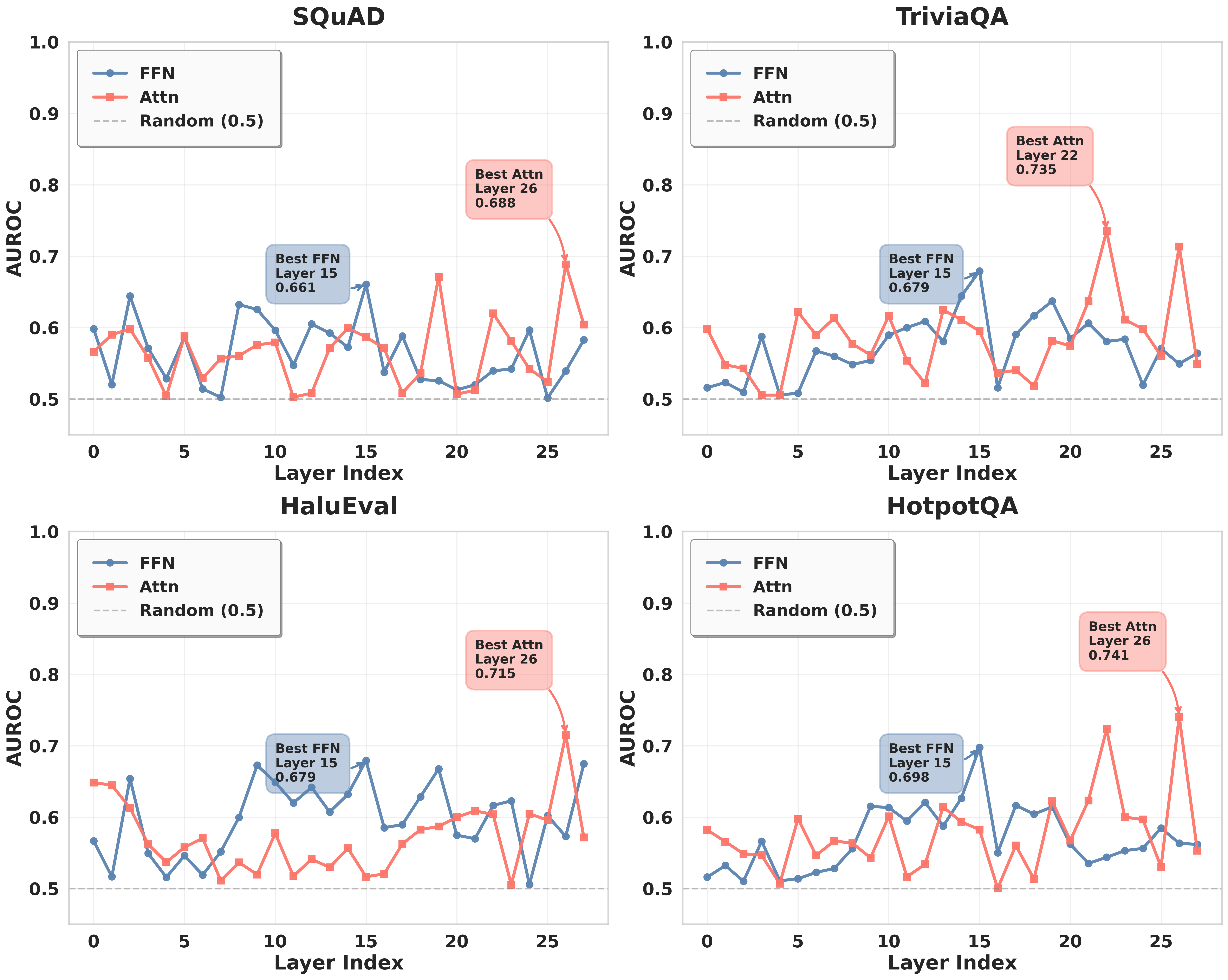}
    \caption{Layer-wise AUROC analysis for Qwen-2.5-7B-Instruct across four datasets: SQuAD (top-left), TriviaQA (top-right), HaluEval (bottom-left), and HotpotQA (bottom-right). Blue curves represent FFN modules; red curves represent Attention modules. Shaded regions indicate $\pm 1$ standard deviation across 5-fold cross-validation. The horizontal dashed line at 0.5 represents random performance.}
    \label{fig:qwen_layer_auroc}
\end{figure*}

Figure~\ref{fig:qwen_layer_auroc} presents the layer-wise AUROC analysis for Qwen-2.5-7B-Instruct. Key observations include:

\paragraph{Dataset-Specific Peak Patterns:}
\begin{itemize}
    \item \textbf{SQuAD (context-rich):} FFN peaks at Layer 15 (AUROC: 0.661), while Attention shows strong performance in late layers, peaking at Layer 26 (AUROC: 0.688).
    \item \textbf{TriviaQA (context-free):} Attention shows a pronounced peak at Layer 22 (AUROC: 0.735), whereas FFN peaks earlier at Layer 15 (AUROC: 0.679).
    \item \textbf{HaluEval (context-rich):} FFN exhibits strong middle-layer performance peaking at Layer 15 (AUROC: 0.679), while Attention outperforms in the final stages, peaking at Layer 26 (AUROC: 0.715).
    \item \textbf{HotpotQA (multi-hop):} FFN peaks at Layer 15 (AUROC: 0.698), while Attention reaches the highest discriminative power at Layer 26 (AUROC: 0.741).
\end{itemize}

\paragraph{Model-Specific Characteristics:}
Qwen-2.5-7B shows a consistent temporal pattern where FFN contributions peak in the middle layers (Layer 15), while Attention contributions tend to dominate in the late layers (Layers 22-26). The peak AUROC values generally range from 0.66 to 0.74.

\subsection{LLaMA-3.1-8B-Instruct}

\begin{figure*}[t]
    \centering
    \includegraphics[width=0.95\textwidth]{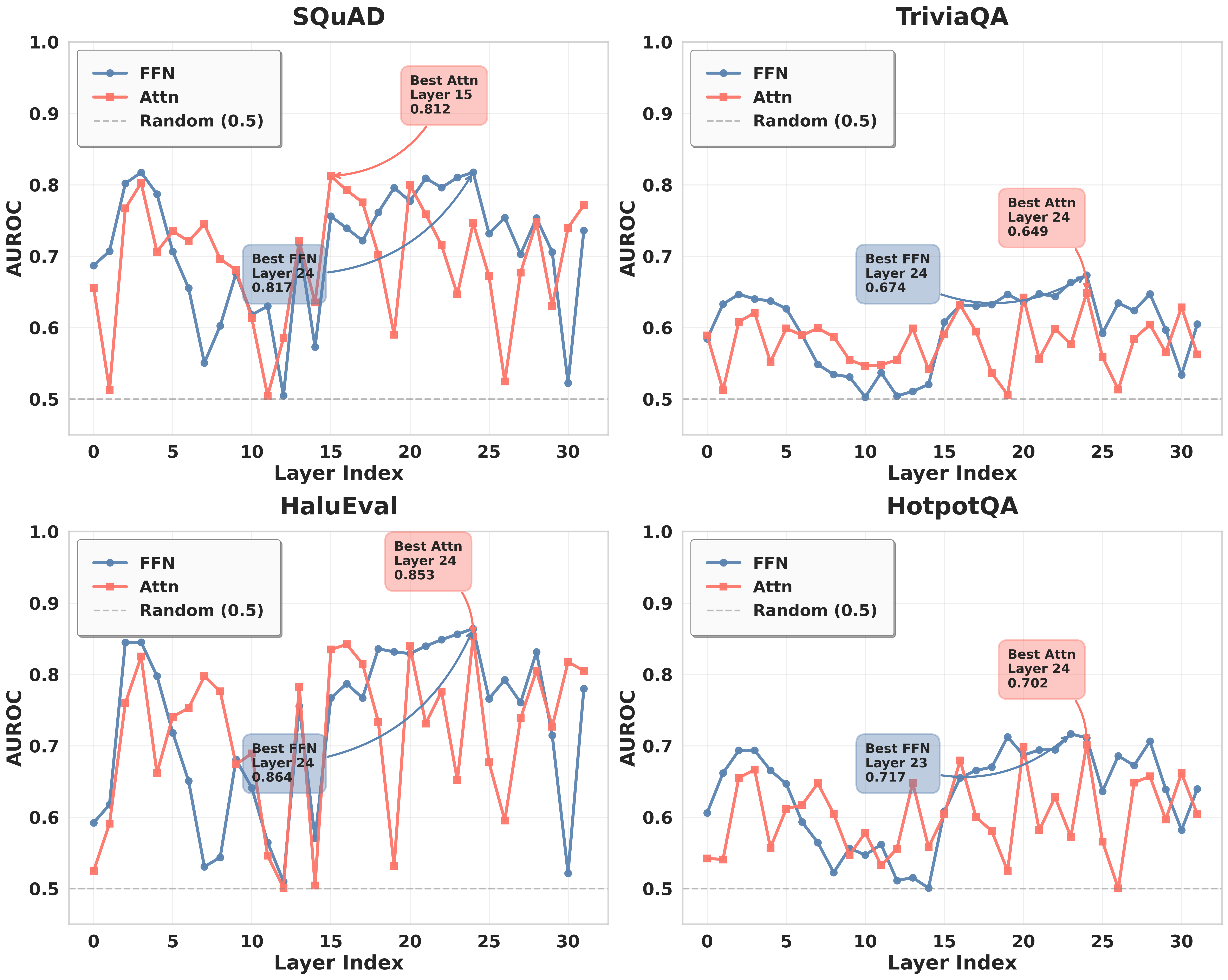}
    \caption{Layer-wise AUROC analysis for LLaMA-3.1-8B-Instruct across four datasets: SQuAD (top-left), TriviaQA (top-right), HaluEval (bottom-left), and HotpotQA (bottom-right). Blue curves represent FFN modules; red curves represent Attention modules. Shaded regions indicate $\pm 1$ standard deviation across 5-fold cross-validation.}
    \label{fig:llama_layer_auroc}
\end{figure*}

Figure~\ref{fig:llama_layer_auroc} presents the layer-wise AUROC analysis for LLaMA-3.1-8B-Instruct. Key observations include:

\paragraph{Dataset-Specific Peak Patterns:}
\begin{itemize}
    \item \textbf{SQuAD:} Attention peaks at Layer 15 (AUROC: 0.812), with FFN showing consistent performance across layers 10-25, peaking at Layer 24 (AUROC: 0.817).
    \item \textbf{TriviaQA:} FFN exhibits the strongest performance at Layer 24 (AUROC: 0.674), while Attention peaks at Layer 24 (AUROC: 0.649).
    \item \textbf{HaluEval:} FFN dominates with a peak at Layer 24 (AUROC: 0.864), while Attention peaks at Layer 24 (AUROC: 0.833), demonstrating clear temporal separation of module contributions.
    \item \textbf{HotpotQA:} Both modules show strong late-layer performance, with FFN peaking at Layer 23 (AUROC: 0.717) and Attention at Layer 24 (AUROC: 0.702).
\end{itemize}

\paragraph{Model-Specific Characteristics:}
LLaMA-3.1-8B achieves the highest peak AUROC values across all models, particularly in FFN modules. The model exhibits pronounced late-layer discriminative power, suggesting that its architecture emphasizes final-stage knowledge integration and projection for uncertainty-related signals.

\subsection{Gemma-2-9B-IT}

\begin{figure*}[t]
    \centering
    \includegraphics[width=0.95\textwidth]{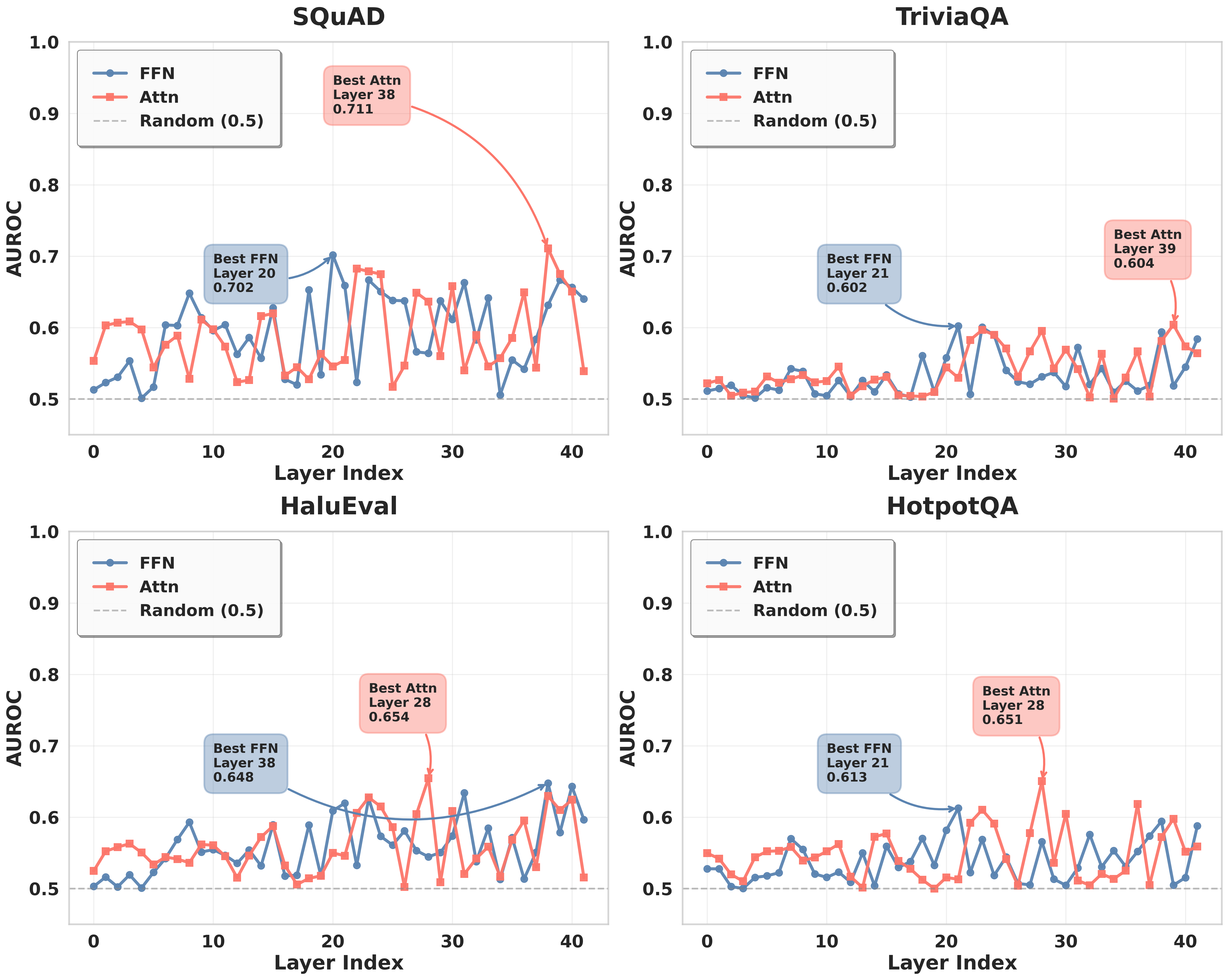}
    \caption{Layer-wise AUROC analysis for Gemma-2-9B-IT across four datasets: SQuAD (top-left), TriviaQA (top-right), HaluEval (bottom-left), and HotpotQA (bottom-right). Blue curves represent FFN modules; red curves represent Attention modules. Shaded regions indicate $\pm 1$ standard deviation across 5-fold cross-validation.}
    \label{fig:gemma_layer_auroc}
\end{figure*}

Figure~\ref{fig:gemma_layer_auroc} presents the layer-wise AUROC analysis for Gemma-2-9B-IT. Key observations include:

\paragraph{Dataset-Specific Peak Patterns:}
\begin{itemize}
    \item \textbf{SQuAD:} Attention shows a strong late-layer peak at Layer 38 (AUROC: 0.711), while FFN peaks earlier at Layer 20 (AUROC: 0.702).
    \item \textbf{TriviaQA:} Both modules show moderate performance, with Attention peaking at Layer 39 (AUROC: 0.604) and FFN at Layer 21 (AUROC: 0.602).
    \item \textbf{HaluEval:} Attention exhibits a peak at Layer 28 (AUROC: 0.654), while FFN shows a late peak at Layer 38 (AUROC: 0.648).
    \item \textbf{HotpotQA:} Attention achieves the highest performance at Layer 28 (AUROC: 0.651), with FFN peaking at Layer 21 (AUROC: 0.613).
\end{itemize}

\paragraph{Model-Specific Characteristics:}
Gemma-2-9B exhibits distinct behavior compared to Qwen and LLaMA, with Attention often outperforming FFN in peak AUROC values across datasets. The model generally shows lower overall AUROC scores (mostly in the 0.60-0.71 range) compared to LLaMA-3.1-8B, suggesting that uncertainty signals in Gemma might be more distributed or harder to isolate using single-layer restoration scores.

\subsection{Cross-Model Summary}

Table~\ref{tab:peak_auroc_summary} summarizes the peak AUROC values for each model-dataset combination, highlighting the best-performing layer and module.

\begin{table*}[t]
\centering
\small
\begin{tabular}{llccc}
\toprule
\textbf{Model} & \textbf{Dataset} & \textbf{Best Module} & \textbf{Best Layer} & \textbf{Peak AUROC} \\
\midrule
\multirow{4}{*}{Qwen-2.5-7B} & SQuAD & Attention & 26 & 0.69 \\
 & TriviaQA & Attention & 22 & 0.74 \\
 & HaluEval & Attention & 26 & 0.72 \\
 & HotpotQA & Attention & 26 & 0.74 \\
\midrule
\multirow{4}{*}{LLaMA-3.1-8B} & SQuAD & FFN & 24 & 0.82 \\
 & TriviaQA & FFN & 24 & 0.67 \\
 & HaluEval & FFN & 24 & 0.86 \\
 & HotpotQA & FFN & 23 & 0.72 \\
\midrule
\multirow{4}{*}{Gemma-2-9B} & SQuAD & Attention & 38 & 0.71 \\
 & TriviaQA & Attention & 39 & 0.60 \\
 & HaluEval & Attention & 28 & 0.65 \\
 & HotpotQA & Attention & 28 & 0.65 \\
\bottomrule
\end{tabular}
\caption{Summary of peak layer-wise AUROC values across all models and datasets. LLaMA-3.1-8B achieves the highest overall peak (0.86) on HaluEval using FFN restoration scores.}
\label{tab:peak_auroc_summary}
\end{table*}

\paragraph{Universal Findings:}

\begin{enumerate}
    \item \textbf{Model-Dependent Module Dominance:} Contrary to a single universal rule, the dominant uncertainty signal varies by architecture. Qwen-2.5 and Gemma-2 consistently achieve peak performance using \textbf{Attention} restoration scores (dominating 8/8 cases combined), whereas LLaMA-3.1 relies exclusively on \textbf{FFN} signals (4/4 cases). This suggests that different architectures may localize failure modes in different components (contextual routing vs. parametric memory).
    
    \item \textbf{Late-Layer Information Concentration:} Across all models, the most discriminative signals are concentrated in the deep layers (Layers 22-26 for Qwen/LLaMA, Layers 28-39 for Gemma). This validates that uncertainty is most effectively detected during the final stages of semantic integration and token projection.

    \item \textbf{Performance Hierarchy:} LLaMA-3.1-8B exhibits the strongest internal signatures of correctness, achieving the highest peak AUROC (0.864), followed by Qwen-2.5 (0.741) and Gemma-2 (0.711). This correlates with the general capability of the base models, suggesting that stronger models possess more distinct mechanistic boundaries between correct and incorrect generations.
\end{enumerate}

These results highlight that while the "late-layer" principle is universal, the specific module (Attention vs. FFN) serving as the best uncertainty proxy is architecture-dependent, necessitating the dual-stream approach of our DUD-Probe.

\section{Experimental Details}
\label{appendix:experimental}
\subsection{Computing Infrastructure}
Our experiments were conducted on a server equipped with 4 NVIDIA A100 GPUs (80 GB memory each), running CUDA 12.4 and CentOS Linux 7 (Core). The system specifications are as follows:

\begin{verbatim}
NAME="CentOS Linux"
VERSION="7 (Core)"
ID="centos"
VERSION_ID="7"
PRETTY_NAME="CentOS Linux 7 (Core)"
\end{verbatim}

Across multiple rounds of experiments, the total computational budget amounted to approximately 800-1000 GPU hours.

\subsection{Prompt Templates}
\label{app:prompt_templates}

In our causal tracing experiments, we employ task-specific prompt templates with few-shot examples to ensure consistent evaluation across different datasets. This section details the exact prompts used for each dataset type.

\subsubsection{Context-Rich Tasks (SQuAD \& HaluEval)}

For tasks where answers can be extracted from provided context passages, we use the following template structure:

\begin{tcolorbox}[
    colback=gray!5,
    colframe=gray!75!black,
    boxrule=0.5pt,
    arc=2mm,
    left=3pt,
    right=3pt,
    top=3pt,
    bottom=3pt,
    title={\small\textbf{System Prompt (Qwen/LLaMA models)}},
    fonttitle=\small
]
\small\ttfamily
You are a helpful assistant. Read the context carefully and answer the question truthfully and concisely.

\vspace{0.2em}
[Few-shot Examples]

\vspace{0.2em}
Now answer the following question based on the context provided.
\end{tcolorbox}

\paragraph{Few-shot Example Format:}
\begin{tcolorbox}[
    colback=blue!3,
    colframe=blue!40!black,
    boxrule=0.5pt,
    arc=1mm,
    left=2pt,
    right=2pt,
    top=2pt,
    bottom=2pt
]
\small\ttfamily
Context: [Context Passage]\\
Question: [Question Text]\\
Answer: [Ground Truth Answer]
\end{tcolorbox}

\paragraph{Test Query Format:}
\begin{tcolorbox}[
    colback=green!3,
    colframe=green!40!black,
    boxrule=0.5pt,
    arc=1mm,
    left=2pt,
    right=2pt,
    top=2pt,
    bottom=2pt
]
\small\ttfamily
Context: [Test Context Passage]\\
Question: [Test Question]\\
Answer:
\end{tcolorbox}

\paragraph{Note on Model Variants:}
For Gemma-2-9B, we consolidate the system prompt, few-shot examples, and transition text into a single user message, as this model architecture does not support separate system messages.

\subsubsection{Context-Free Tasks (TriviaQA \& HotpotQA)}

For parametric knowledge tasks without external context, we use a simplified template:

\begin{tcolorbox}[
    colback=gray!5,
    colframe=gray!75!black,
    boxrule=0.5pt,
    arc=2mm,
    left=3pt,
    right=3pt,
    top=3pt,
    bottom=3pt,
    title={\small\textbf{System Prompt (Qwen/LLaMA models)}},
    fonttitle=\small
]
\small\ttfamily
You are a helpful assistant. Answer the question truthfully and concisely.

\vspace{0.2em}
[Few-shot Examples]

\vspace{0.2em}
Now answer the following question.
\end{tcolorbox}

\paragraph{Few-shot Example Format:}
\begin{tcolorbox}[
    colback=blue!3,
    colframe=blue!40!black,
    boxrule=0.5pt,
    arc=1mm,
    left=2pt,
    right=2pt,
    top=2pt,
    bottom=2pt
]
\small\ttfamily
Question: [Question Text]\\
Answer: [Ground Truth Answer]
\end{tcolorbox}

\paragraph{Test Query Format:}
\begin{tcolorbox}[
    colback=green!3,
    colframe=green!40!black,
    boxrule=0.5pt,
    arc=1mm,
    left=2pt,
    right=2pt,
    top=2pt,
    bottom=2pt
]
\small\ttfamily
Question: [Test Question]\\
Answer:
\end{tcolorbox}

\subsubsection{Implementation Details}

\paragraph{Few-shot Configuration:}
We use the first 3 examples from each dataset as few-shot demonstrations, following standard practice in prompt-based evaluation. The remaining samples form the test pool.

\paragraph{Model-Specific Adaptations:}
\begin{itemize}
    \item \textbf{Qwen-2.5 \& LLaMA-3.1:} Use separate \texttt{system} and \texttt{user} roles via the model's chat template API.
    \item \textbf{Gemma-2-9B:} Concatenate all prompt components into a single \texttt{user} message due to architecture constraints.
\end{itemize}

\paragraph{Generation Settings:}
All experiments use the following hyperparameters:
\begin{itemize}
    \item Maximum new tokens: 32
    \item Temperature: Default (greedy decoding)
    \item Padding token: Model-specific EOS token
    \item Noise level: 0.15 for causal tracing
\end{itemize}

\subsubsection{Complete Example}

To illustrate the full prompt structure, we provide a concrete example from the SQuAD dataset:

\begin{tcolorbox}[
    enhanced,
    colback=yellow!5,
    colframe=orange!75!black,
    boxrule=0.8pt,
    arc=2mm,
    title={\textbf{Complete Prompt Example (SQuAD)}},
    fonttitle=\normalsize\bfseries,
    attach boxed title to top left={yshift=-2mm, xshift=3mm},
    boxed title style={colback=orange!75!black}
]
\footnotesize

\textbf{System:}\\
\texttt{You are a helpful assistant. Read the context carefully and answer the question truthfully and concisely.}

\vspace{0.3em}
\textbf{[Few-shot Example 1]}\\
\texttt{Context: The Normans (Norman: Normaunds; French: Normands) were the people who in the 10th and 11th centuries gave their name to Normandy...}\\
\texttt{Question: In what country is Normandy located?}\\
\texttt{Answer: France}

\vspace{0.3em}
\textit{[Examples 2-3 omitted for brevity]}

\vspace{0.3em}
\texttt{Now answer the following question based on the context provided.}

\vspace{0.3em}
\textbf{User:}\\
\texttt{Context: [Test Context]}\\
\texttt{Question: [Test Question]}\\
\texttt{Answer:}
\end{tcolorbox}

\paragraph{Design Rationale:}
This prompt structure balances clarity, consistency, and practical constraints:
\begin{enumerate}
    \item \textbf{Explicit Instructions:} Clear task definition reduces output ambiguity
    \item \textbf{Standardized Format:} Uniform structure across samples minimizes confounding variables
    \item \textbf{Moderate Few-shot:} Three examples provide sufficient task demonstration without excessive context length
\end{enumerate}

\subsection{Datasets}
\label{app:datasets}

For our experiments, we use 10,000 instances from each dataset. Specifically, we use the QA subset of the HaluEval dataset and the standard splits for other datasets. The datasets used in this study are publicly available and adhere to their respective licenses, which allow for academic research and non-commercial use. 

\paragraph{Dataset Details:}
\begin{itemize}
    \item \textbf{HaluEval} \cite{li2023halueval}: A large-scale hallucination evaluation benchmark focusing on QA tasks
    \item \textbf{SQuAD} \cite{rajpurkar2016squad}: Stanford Question Answering Dataset for reading comprehension
    \item \textbf{TriviaQA} \cite{joshi2017triviaqa}: Large-scale reading comprehension dataset with evidence documents
    \item \textbf{HotpotQA} \cite{yang2018hotpotqa}: Multi-hop question answering dataset requiring reasoning over multiple documents
\end{itemize}

Each dataset is split into 80\%-20\% for training and testing. We train and test on each dataset separately, reporting the corresponding results. This experimental setup for baseline methods matches ours exactly.

\subsection{Baseline Methods}
\label{app:baselines}

This section provides a brief overview of several baseline methods for hallucination detection.

\paragraph{PPL (Perplexity):} PPL assesses the likelihood of hallucinations in LLM-generated responses by calculating perplexity \cite{ren2022out}. A higher perplexity value reflects increased uncertainty in the model's output, as hallucinations are often associated with the model's lack of confidence or inaccurate knowledge.

\paragraph{Length-Normalized Entropy (LN-Entropy):} LN-Entropy is designed to quantify sequence-level uncertainty across multiple generations \cite{malinin2020uncertainty}. This metric normalizes entropy with respect to sequence length, facilitating a more precise assessment of the stability and reliability of generated content.

\paragraph{Shifting Attention to Relevance (SAR)}: Proposed by \cite{duan2024shifting}, SAR is a single-pass method that computes uncertainty by focusing only on the ``relevant'' tokens in a generation. It first calculates token-level uncertainty (e.g., entropy) and then re-weights these scores based on each token's relevance to the core answer. This relevance is determined by analyzing the model's internal states (e.g., attention or gradients), allowing the method to disregard uncertainty arising from peripheral or non-essential tokens.

\paragraph{Semantic Entropy (SE)}: Proposed by \cite{kuhn2023semantic}, this method first samples \(M\) candidate answers, groups them into clusters based on semantic equivalence, and then calculates the entropy over the probability distribution of these semantic clusters.

\paragraph{LLM-Check:} LLM-check uses attention mechanism kernel similarity analysis to conduct hallucination detection \cite{zhang2025siren}. By calculating the log-determinant of attention kernel matrices and aggregating across heads, it captures characteristics related to hallucinations.

\paragraph{SAPLMA:} SAPLMA trains a classifier to detect hallucinations using the hidden states of specific layers in LLMs \cite{chen2024inside}. It captures internal signals from these layers to identify when the model is likely to generate hallucinated content.

\paragraph{SEP (Semantic Entropy Probe):} SEP leverages linear probes trained on the hidden states of large language models \cite{kossen2024semantic}. It detects hallucinations by analyzing the semantic entropy of tokens before generation.

\paragraph{ICR (Internal Consistency Ratio):} ICR tracks the dynamic evolution of hidden state updates across layers by measuring the contribution ratio between Attention and FFN modules to the residual stream \cite{zhang2025icr}. The ICR Probe trains a classifier on the layer-wise ICR scores to detect hallucinations, capturing the global update patterns throughout the model's forward pass rather than focusing on isolated layer features.
\subsection{Causal Tracing Probe Training}
\label{app:ct_training}

\paragraph{Input and Output:} 
The Causal Tracing Probe is a multi-layer perceptron (MLP) classifier trained to predict hallucinations using the pooled restoration scores. Given a model generation, the restoration score matrix $\mathbf{R}^{2L \times N}$ (where $L$ is the number of layers and $N$ is the number of generated tokens) is token-wise pooled to obtain a $1 \times 2L$ vector, which serves as the input to the probe. Specifically, dimensions $1$ to $L$ contain the pooled Attention restoration scores, and dimensions $L+1$ to $2L$ contain the pooled FFN restoration scores. The label for each instance is derived from annotated hallucination datasets.

\paragraph{Model Architecture:} 
The probe consists of four fully connected layers, with batch normalization applied after each layer to stabilize training. The architecture follows the configuration $(2L, 128, 64, 32, 1)$, where $2L$ is the input dimension (twice the number of layers due to separate Attention and FFN scores). For models with approximately $L = 28$ layers, the total number of parameters remains below 16K, maintaining computational efficiency. ReLU activation functions are applied between layers, and the final layer uses a sigmoid activation to output a hallucination probability score.

\paragraph{Details of Training:}
\begin{itemize}
    \item The model is trained using the binary cross-entropy loss with the Adam optimizer (learning rate = $5 \times 10^{-4}$).
    \item We employ learning rate scheduling with a cosine annealing strategy.
    \item Training is performed for 50 epochs with early stopping based on validation AUROC.
    \item Batch size is set to 256, and we apply dropout (rate = 0.3) for regularization.
    \item The dataset is split into 80\% training and 20\% testing sets.
    \item We perform 5-fold cross-validation and report the average performance.
\end{itemize}

\begin{algorithm*}[h]
\caption{Decoupled Feature Extraction via Causal Tracing}
\label{alg:extraction}
\begin{algorithmic}[1]
\Require LLM $\mathcal{M}$, Input prompt $x$, Answer tokens $y$, Number of layers $L$, Noise level $\nu$
\Ensure Feature vector $\mathbf{v} \in \mathbb{R}^{2L}$

\State \textbf{Step 1: Clean Run}
\State Run $\mathcal{M}(x)$ to get clean probs $P_{\text{clean}}$ and cache activations $\{\mathbf{m}^{(l)}, \mathbf{a}^{(l)}\}_{l=1}^L$

\State \textbf{Step 2: Corrupted Run}
\State $x^* \leftarrow x + \epsilon, \text{ where } \epsilon \sim \mathcal{N}(0, \nu)$ \Comment{Inject noise to embeddings}
\State Run $\mathcal{M}(x^*)$ to get corrupted probs $P_{\text{corr}}$
\State $\text{TE} \leftarrow P_{\text{clean}} - P_{\text{corr}}$ \Comment{Calculate Total Effect}

\State \textbf{Step 3: Decoupled Restoration (Patching)}
\State Initialize score lists $S_{\text{ffn}} \leftarrow [], S_{\text{attn}} \leftarrow []$
\For{$l = 1$ to $L$}
    \State \textit{// Patch FFN Module}
    \State Restore $\mathbf{m}^{(l)}$ into $\mathcal{M}(x^*)$
    \State $P_{\text{restored}}^{\text{ffn}} \leftarrow$ Probs from patched run
    \State $s_{\text{ffn}} \leftarrow (P_{\text{restored}}^{\text{ffn}} - P_{\text{corr}}) / \text{TE}$
    \State Append $s_{\text{ffn}}$ to $S_{\text{ffn}}$
    
    \State \textit{// Patch Attention Module}
    \State Restore $\mathbf{a}^{(l)}$ into $\mathcal{M}(x^*)$
    \State $P_{\text{restored}}^{\text{attn}} \leftarrow$ Probs from patched run
    \State $s_{\text{attn}} \leftarrow (P_{\text{restored}}^{\text{attn}} - P_{\text{corr}}) / \text{TE}$
    \State Append $s_{\text{attn}}$ to $S_{\text{attn}}$
\EndFor

\State \textbf{Return} $\mathbf{v} \leftarrow \text{Concat}(S_{\text{ffn}}, S_{\text{attn}})$
\end{algorithmic}
\end{algorithm*}
\section{Implementation Details of DUD-Probe}
\label{app:implementation_details}

\subsection{Probe Architecture and Training}
\label{app:probe_arch}

\begin{algorithm*}[h]
\caption{DUD-Probe Training and Inference}
\label{alg:training}
\begin{algorithmic}[1]
\Require Training dataset $\mathcal{D} = \{(\mathbf{v}_i, y_i)\}_{i=1}^N$, Learning rate $\eta$, Batch size $B$
\Ensure Trained probe parameters $\theta$

\Procedure{Train}{$\mathcal{D}$}
    \State Initialize MLP weights $\theta$ (Kaiming Uniform)
    \For{epoch $= 1$ to $E$}
        \For{batch $(\mathbf{V}, \mathbf{Y})$ in $\mathcal{D}$}
            \State $\hat{\mathbf{Y}} \leftarrow \sigma(\text{MLP}_\theta(\mathbf{V}))$ \Comment{Forward pass}
            \State $\mathcal{L} \leftarrow \text{BCELoss}(\hat{\mathbf{Y}}, \mathbf{Y})$
            \State $\theta \leftarrow \theta - \eta \nabla_\theta \mathcal{L}$ \Comment{Adam update}
        \EndFor
        \State Update $\eta$ via Scheduler based on val loss
    \EndFor
    \State \Return $\theta$
\EndProcedure

\Statex

\Procedure{Inference}{Prompt $x$, Answer $y$, $\mathcal{M}$, $\theta$}
    \State $\mathbf{v} \leftarrow \Call{FeatureExtraction}{x, y, \mathcal{M}}$ \Comment{See Alg. \ref{alg:extraction}}
    \State $\text{uncertainty} \leftarrow \sigma(\text{MLP}_\theta(\mathbf{v}))$
    \State \Return $\text{uncertainty}$
\EndProcedure
\end{algorithmic}
\end{algorithm*}
\paragraph{Input Representation.} 
Unlike previous methods that aggregate residual stream information into a single scalar per layer, the DUD-Probe operates on a high-dimensional feature space that preserves the independent contributions of different modules. For a model with $L$ layers, the input is a concatenated vector $\mathbf{v} \in \mathbb{R}^{2L}$, constructed as:
\begin{equation}
    \mathbf{v} = \text{Concat}\left( [\mathcal{S}_{\text{ffn}}^{(1)}, \dots, \mathcal{S}_{\text{ffn}}^{(L)}], [\mathcal{S}_{\text{attn}}^{(1)}, \dots, \mathcal{S}_{\text{attn}}^{(L)}] \right)
\end{equation}
where $\mathcal{S}_{\text{ffn}}^{(l)}$ and $\mathcal{S}_{\text{attn}}^{(l)}$ are the decoupled causal restoration scores for the FFN and Attention modules at layer $l$, respectively.

\paragraph{Model Configuration.} 
We employ a Multi-Layer Perceptron (MLP) to capture the non-linear interactions between memory (FFN) and context (Attention) signals. Based on empirical tuning, we adopt a 4-layer architecture with the dimension progression: $2L \to 128 \to 64 \to 32 \to 1$.
To ensure training stability and prevent overfitting on the probe dataset, we apply the following regularization techniques:
\begin{itemize}
    \item \textbf{Batch Normalization:} Applied after each linear transformation to standardize layer inputs.
    \item \textbf{Activation Function:} We use \textbf{LeakyReLU} (negative slope $\alpha=0.01$) for all hidden layers to handle potential negative values in the causal scores, while the final output uses a \textbf{Sigmoid} function to produce a probability $p \in [0, 1]$.
    \item \textbf{Dropout:} A dropout rate of $p=0.3$ is applied after the activation function of each hidden layer.
    \item \textbf{Initialization:} Linear layers are initialized using Kaiming Uniform initialization.
\end{itemize}

\paragraph{Label Generation.}
To train the probe, we generate binary labels ($y \in \{0, 1\}$) for the model's responses. We use the ROUGE-L F-measure to compare the generated response with the ground truth answer. Following standard practices in hallucination detection benchmarks (e.g., HaluEval), a response is labeled as \textit{Faithful} ($y=0$) if the ROUGE-L score $\ge 0.5$, and \textit{Hallucinated} ($y=1$) otherwise.

\paragraph{Training Hyperparameters.}
The probe is trained using the Binary Cross-Entropy (BCE) loss. We use the Adam optimizer with an initial learning rate of $5 \times 10^{-4}$. We employ a learning rate scheduler (`ReduceLROnPlateau`) that reduces the learning rate by a factor of 0.5 if the validation loss does not improve for 5 consecutive epochs. The training is conducted with a batch size of 32 for up to 50 epochs, with early stopping triggered if validation performance stagnates.

\section{Algorithms}

We summarize the procedures for extracting the decoupled dynamic features (Algorithm \ref{alg:extraction}) and training the DUD-Probe (Algorithm \ref{alg:training}).

\section{Additional Experimental Results}
\label{app:additional_metrics}

To further validate the reliability and practical utility of our uncertainty estimation, we present additional evaluation metrics on the Qwen2.5-7B model: \textbf{Expected Calibration Error (ECE)} and \textbf{Prediction Rejection Ratio (PRR)}.

\subsection{Calibration Performance (ECE)}

ECE measures the alignment between the model's predicted confidence and its actual accuracy. Lower ECE values indicate better calibration. As shown in Table~\ref{tab:ece_results}, DUD-Probe achieves significantly lower ECE scores across all datasets compared to baselines. Notably, on HotpotQA, our method reaches an ECE of \textbf{0.0039}, which is orders of magnitude better than SEP (0.2879) and SAPLMA (0.1675). This indicates that the uncertainty scores produced by our probe can be interpreted as reliable probabilities of correctness, which is crucial for downstream decision-making.

\begin{table*}[h]
\centering
\setlength{\tabcolsep}{6pt}
\begin{tabular}{l|cccc}
\toprule
\textbf{Method} & \textbf{HaluEval} & \textbf{SQuAD} & \textbf{HotpotQA} & \textbf{TriviaQA} \\ \midrule
SEP & 0.1431 & 0.1537 & 0.0979 & 0.1019 \\
SAPLMA & 0.0997 & 0.1578 & 0.0675 & 0.0435 \\
ICR Probe & 0.0798 & 0.0314 & 0.0138 & 0.0662 \\
\textbf{DUD-Probe} & \textbf{0.0425} & \textbf{0.0173} & \textbf{0.0039} & \textbf{0.0289} \\ \bottomrule
\end{tabular}
\caption{ECE performance on Qwen2.5-7B (ROUGE-L $>$ 0.5). Lower values indicate better calibration.}
\label{tab:ece_results}
\end{table*}

\subsection{Rejection Efficiency (PRR)}

PRR evaluates how effectively an uncertainty metric can filter out incorrect predictions. Higher PRR values indicate that rejecting samples with high uncertainty leads to a greater improvement in the overall accuracy of the remaining set. Table~\ref{tab:prr_results} demonstrates that DUD-Probe consistently achieves the highest PRR scores. For example, on HaluEval, our method achieves a PRR of \textbf{0.6346}, significantly outperforming ICR Probe (0.5003). This confirms that our decoupled features provide a highly discriminative signal for separating correct from incorrect generations.

\begin{table*}[h]
\centering
\setlength{\tabcolsep}{6pt}
\begin{tabular}{l|cccc}
\toprule
\textbf{Method} & \textbf{HaluEval} & \textbf{SQuAD} & \textbf{HotpotQA} & \textbf{TriviaQA} \\ \midrule
SEP & 0.3927 & 0.3578 & 0.3243 & 0.3696 \\
SAPLMA & 0.4414 & 0.4675 & 0.4373 & 0.4459 \\
ICR Probe & 0.4989 & 0.5204 & 0.4361 & 0.5435 \\
\textbf{DUD-Probe} & \textbf{0.6346} & \textbf{0.5669} & \textbf{0.5187} & \textbf{0.5945} \\ \bottomrule
\end{tabular}
\caption{PRR performance on Qwen2.5-7B (ROUGE-L $>$ 0.5). Higher values indicate better rejection efficiency.}
\label{tab:prr_results}
\end{table*}

\section{Sensitivity to Labeling Criteria}
\label{app:threshold_analysis}

The definition of "correctness" in generation tasks can be subjective. To ensure that our performance gains are not artifacts of a specific labeling threshold (i.e., ROUGE-L $>$ 0.5 used in the main text), we evaluate the robustness of the DUD-Probe on Qwen2.5-7B under varying strictness levels and alternative metrics.

\subsection{Alternative Metrics: Exact Match and BERTScore}

We further evaluate performance using \textbf{Exact Match (EM)}, which requires the generation to be identical to the ground truth, and \textbf{BERTScore-F1 ($>$ 0.95)}, which measures semantic similarity.

Results in Table~\ref{tab:metric_sensitivity} demonstrate the superior adaptability of DUD-Probe.
\begin{itemize}
    \item \textbf{Exact Match:} DUD-Probe achieves remarkable performance on SQuAD (\textbf{0.8801}) and HotpotQA (\textbf{0.8717}), significantly surpassing the ICR Probe. This suggests that decoupled traces are highly sensitive to the precise retrieval of correct entities required for exact matching.
    \item \textbf{BERTScore:} In semantic evaluation, our method maintains its lead (e.g., \textbf{0.8969} on HotpotQA), confirming that the probe captures semantic correctness rather than just lexical overlap.
\end{itemize}

\begin{table*}[h]
\centering
\setlength{\tabcolsep}{5pt}
\begin{tabular}{c|l|cccc}
\toprule
\textbf{Metric} & \textbf{Method} & \textbf{HaluEval} & \textbf{SQuAD} & \textbf{HotpotQA} & \textbf{TriviaQA} \\ \midrule
\multirow{4}{*}{\shortstack{Exact\\Match}} 
 & SEP & 0.6544 & 0.6038 & 0.6359 & 0.7754 \\
 & SAPLMA & 0.7341 & 0.7267 & 0.7148 & 0.7749 \\
 & ICR Probe & 0.7554 & 0.7543 & 0.7497 & 0.7903 \\
 & \textbf{DUD-Probe} & \textbf{0.8204} & \textbf{0.8801} & \textbf{0.8717} & \textbf{0.8461} \\ \midrule
\multirow{4}{*}{\shortstack{BERTScore\\$>0.95$}} 
 & SEP & 0.6643 & 0.6111 & 0.6799 & 0.7034 \\
 & SAPLMA & 0.7267 & 0.7313 & 0.7524 & 0.7348 \\
 & ICR Probe & 0.7347 & 0.7432 & 0.7781 & 0.7438 \\
 & \textbf{DUD-Probe} & \textbf{0.8111} & \textbf{0.8541} & \textbf{0.8969} & \textbf{0.8249} \\ \bottomrule
\end{tabular}
\caption{AUROC performance using Exact Match and BERTScore labeling on Qwen2.5-7B.}
\label{tab:metric_sensitivity}
\end{table*}

\section{Model Size Comparison}
\label{app:model_size}

To evaluate the robustness of our method across different model scales, we conduct experiments on two variants of the Qwen2.5 family: Qwen2.5-3B and Qwen2.5-14B. Table~\ref{tab:model_size_comparison} presents the AUROC performance of DUD-Probe and baseline methods across both model sizes.

\begin{table*}[h]
\centering
\setlength{\tabcolsep}{8pt}
\begin{tabular}{@{}llcccc@{}}
\toprule
\textbf{Model} & \textbf{Method} & \textbf{HaluEval} & \textbf{SQuAD} & \textbf{HotpotQA} & \textbf{TriviaQA} \\ \midrule
\multirow{4}{*}{Qwen2.5-3B} 
 & SAPLMA & 0.7638 & 0.7015 & 0.7547 & 0.7424 \\
 & SEP & 0.6589 & 0.6606 & 0.6929 & 0.6745 \\
 & ICR Probe & 0.8017 & 0.7484 & 0.7995 & 0.7731 \\
 & \textbf{DUD-Probe} & \textbf{0.9212} & \textbf{0.7894} & \textbf{0.8689} & \textbf{0.8475} \\ \midrule
\multirow{4}{*}{Qwen2.5-14B} 
 & SAPLMA & 0.7527 & 0.7088 & 0.7541 & 0.7559 \\
 & SEP & 0.6961 & 0.6955 & 0.6974 & 0.7003 \\
 & ICR Probe & 0.8121 & 0.7432 & 0.7951 & 0.7546 \\
 & \textbf{DUD-Probe} & \textbf{0.8867} & \textbf{0.7911} & \textbf{0.8261} & \textbf{0.9209} \\ \bottomrule
\end{tabular}
\caption{AUROC comparison across different model sizes (ROUGE-L $>$ 0.5). DUD-Probe consistently outperforms baselines regardless of model scale.}
\label{tab:model_size_comparison}
\end{table*}

\paragraph{Key Observations:}
\begin{enumerate}
    \item \textbf{Consistent Superiority:} DUD-Probe achieves the highest AUROC across both model sizes and all datasets, confirming that the effectiveness of decoupled causal tracing is independent of model capacity.
    \item \textbf{Scale Robustness:} While larger models (14B) generally improve baseline performance slightly, the relative advantage of DUD-Probe over ICR Probe remains substantial (e.g., +7.5\% on HaluEval for 14B, +12.0\% for 3B), indicating that our method scales well.
    \item \textbf{Small Model Gains:} On the smaller Qwen2.5-3B model, DUD-Probe demonstrates particularly strong improvements over aggregated methods (ICR: 0.8017 $\to$ DUD: 0.9212 on HaluEval), suggesting that explicit decoupling is especially valuable when model capacity is limited and internal conflicts are more pronounced.
\end{enumerate}

\section{Cross-Model Generalization Analysis}
\label{app:cross_model_generalization}

To demonstrate the universality of our proposed framework, we extend the cross-dataset generalization analysis to two additional state-of-the-art open-source model families: \textbf{Qwen2.5-7B} and \textbf{Gemma-2-9B}. This ensures that the effectiveness of DUD-Probe is not limited to a specific architecture but is applicable across different Transformer implementations.

\textbf{Results on Qwen2.5.} Table~\ref{tab:qwen_generalization} presents the transferability results for Qwen2.5-7B. The model exhibits exceptional stability, with in-domain AUROC scores consistently exceeding 0.82. Notably, the transfer performance is remarkably high; for instance, a probe trained on \textit{HotpotQA} achieves an AUROC of 0.8294 when tested on \textit{TriviaQA}, a drop of less than 2\% compared to the in-domain baseline. This suggests that Qwen2.5's internal uncertainty signatures are highly consistent across different knowledge-intensive tasks.

\textbf{Results on Gemma-2.} Table~\ref{tab:gemma_generalization} details the performance on Gemma-2-9B. Despite architectural differences (e.g., sliding window attention, logit capping), DUD-Probe maintains robust generalization capabilities. The method achieves an average in-domain AUROC of approximately 0.84. While the transfer from \textit{SQuAD} to other datasets shows a slight variance, the overall performance remains significantly superior to random guessing and competitive with in-domain training, further validating the robustness of decoupled dynamics as a generalized uncertainty proxy.

These findings, combined with the Llama-3 results in the main text, confirm that the mechanistic conflict between parametric memory and contextual processing is a fundamental indicator of uncertainty invariant across diverse LLM architectures.

\begin{table*}[h]
\centering
\caption{Cross-dataset generalization evaluation for \textbf{Qwen2.5-7B} (DUD-Probe). Each cell shows the AUROC when the probe is trained on the row dataset and evaluated on the column dataset. Diagonal entries (in-domain performance) are bolded.}
\label{tab:qwen_generalization}
\begin{tabular}{cc|cccc}
\toprule
\multicolumn{2}{c|}{\textbf{Qwen2.5}} & \multicolumn{4}{c}{\textbf{Test}} \\
\cmidrule(lr){3-6}
& & \textbf{HaluEval} & \textbf{SQuAD} & \textbf{HotpotQA} & \textbf{TriviaQA} \\
\midrule
\multirow{4}{*}{\textbf{Train}} 
& HaluEval & \textbf{0.8642} & 0.8059 & 0.8326 & 0.7685 \\
& SQuAD & 0.8270 & \textbf{0.8204} & 0.8275 & 0.7652 \\
& HotpotQA & 0.7860 & 0.7533 & \textbf{0.8435} & 0.7870 \\
& TriviaQA & 0.7543 & 0.7183 & 0.8294 & \textbf{0.8252} \\
\bottomrule
\end{tabular}
\end{table*}

\begin{table*}[h]
\centering
\caption{Cross-dataset generalization evaluation for \textbf{Gemma-2-9B} (DUD-Probe). Each cell shows the AUROC when the probe is trained on the row dataset and evaluated on the column dataset. Diagonal entries (in-domain performance) are bolded.}
\label{tab:gemma_generalization}
\begin{tabular}{cc|cccc}
\toprule
\multicolumn{2}{c|}{\textbf{Gemma-2}} & \multicolumn{4}{c}{\textbf{Test}} \\
\cmidrule(lr){3-6}
& & \textbf{HaluEval} & \textbf{SQuAD} & \textbf{HotpotQA} & \textbf{TriviaQA} \\
\midrule
\multirow{4}{*}{\textbf{Train}} 
& HaluEval & \textbf{0.8747} & 0.7941 & 0.7668 & 0.7658 \\
& SQuAD & 0.7567 & \textbf{0.8218} & 0.7436 & 0.7555 \\
& HotpotQA & 0.7759 & 0.7749 & \textbf{0.8597} & 0.7632 \\
& TriviaQA & 0.7992 & 0.7625 & 0.7741 & \textbf{0.8199} \\
\bottomrule
\end{tabular}
\end{table*}
\end{document}